\pdfoutput=1

\documentclass[11pt]{article}

\usepackage[]{acl}

\usepackage{times}
\usepackage{latexsym}

\usepackage[T1]{fontenc}

\usepackage[utf8]{inputenc}

\usepackage{microtype}

\usepackage{amsfonts}
\usepackage{caption}
\usepackage{subcaption}
\usepackage{enumitem}
\usepackage{wrapfig}
\usepackage{multirow}
\usepackage{booktabs}
\usepackage{graphicx}
\usepackage{amsmath}
\usepackage{amsthm}
\usepackage{booktabs}
\usepackage{algorithm}
\usepackage{algorithmic}

\title{Improved Training of Mixture-of-Experts Language GANs}

\author{Yekun Chai \\
  Baidu \\
  \texttt{chaiyekun@baidu.com}  \\\And
  Qiyue Yin \and Junge Zhang\thanks{ \, Corresponding author.} \\
  Institute of Automation, CAS \\
\texttt{\{qyyin,jgzhang\}@nlpr.ia.ac.cn} \\}

\begin{document}
\maketitle
\begin{abstract}
Despite the dramatic success in image generation, Generative Adversarial Networks (GANs) still face great challenges in synthesizing sequences of discrete elements, in particular human language. The difficulty in generator training arises from the limited representation capacity and uninformative learning signals obtained from the discriminator. In this work, we (1) first empirically show that the mixture-of-experts approach is able to enhance the representation capacity of the generator for language GANs and (2) harness the Feature Statistics Alignment (FSA) paradigm to render fine-grained learning signals to advance the generator training.  Specifically, FSA forces the mean statistics of the distribution of fake data to approach that of real samples as close as possible in the finite-dimensional feature space. Empirical study on synthetic and real benchmarks shows the superior performance in quantitative evaluation and demonstrates the effectiveness of our approach to adversarial text generation. 
\end{abstract}

\section{Introduction}
\label{sec:intro}
Unsupervised sequence generation is the cornerstone for a plethora of applications, such as dialogue generation~\cite{li2017adversarial}. The most common approach to autoregressive sequence modeling is maximizing the likelihood of each token in the sequence given the previous partial observation. However, using maximum likelihood estimation (MLE) is inherently prone to the exposure bias problem~\cite{bengio2015scheduled}, which results from the discrepancy between the training and inference stage: the generator predicts the next token conditioned on its previously generated ones during inference but on its prefix ground-truth tokens during training, yielding accumulative mismatch along with the increment of generated sequence length.

Generative Adversarial Networks  (GANs) ~\cite{Goodfellow2014GenerativeAN} can serve as an alternative to models trained with MLE, which have achieved promising results in generating sequences of discrete elements, in particular, language sequences~\cite{Kusner2016GANSFS,Yu2017SeqGANSG,Lin2017AdversarialRF,Guo2018LongTG,fedus2018maskgan,Nie2019RelGANRG,dAutume2019TrainingLG,Zhou2020SelfAdversarialLW,Scialom2020ColdGANsTL}. GANs consist of two competing networks: a discriminator that is trained to distinguish the generated samples from real data, and a generator that aims to generate high-quality samples to fool the discriminator. 


Although language GANs have succeeded in avoiding exposure bias issues, the limited representation capacity of the generator still precludes them from covering complex patterns of the real data distribution and thus deteriorates the quality and diversity of generated samples~\cite{Nie2019RelGANRG}. There has been a recent move towards enhancing the expressive power of generators in language GANs~\cite{Nie2019RelGANRG,liu2020catgan,Scialom2020ColdGANsTL} by leveraging advanced building blocks in the generator architecture, such as Relational Recurrent Neural Networks~\cite{santoro2018relational} or pretrained Transformers~\cite{Scialom2020ColdGANsTL}. 

Inspired by the preeminence of mixture-of-experts approaches in image GANs~\cite{hoang2018mgan,ghosh2018multi} and language generation tasks, such as machine translation~\cite{bi2019multi}, we propose to adopt a mixture-of-experts generator to enhance the representation capacity of the proposed model. Generally, the generator consists of multiple cooperative experts who attempt to jointly generate high-quality sentences through their interaction to compete with the discriminator. This is orthogonal to the aforementioned improvements on generators in language GANs.

Meanwhile, lacking sufficient learning signals is another reason that hampers language GAN's performance~\cite{Lin2017AdversarialRF, chai-etal-2021-counter-contrastive}. \citet{Lin2017AdversarialRF} claimed that the binary classification in the discriminator network limits the learning capacity of tasks because the diversity and richness are plagued by the degenerated distribution. RankGAN~\cite{Lin2017AdversarialRF} replaced the binary classifier with a pairwise feature ranker by comparing the similarities between sample features in the latent space. SAL~\cite{Zhou2020SelfAdversarialLW} classified the encoded features of constructed pairwise training examples into three categories, \emph{i.e.}, better / worse / indistinguishable.

To further promote the generator's training by enriching the learning signals aside from the discriminator, we use an auxiliary encoder to extract the latent embeddings from real and fake data distributions and endow the generator with additional updating feedback apart from that yielded by the discriminator. Specifically, we leverage the Feature Statistics Alignment (FSA) paradigm to embed such latent feature representations in the finite-dimensional feature space and force the distribution of generated samples to approach the real data distribution by minimizing the distance between their respective feature representation centroids. Intuitively, matching the mean feature representations of fake and real instances could make the two data distributions closer. These intuitive advantages are also borne out in our ablation analysis.

Overall, we introduce a GAN architecture with a mixture-of-experts generator and Feature Statistics Alignment method, termed MoEGAN, for sequence generation. Our experimental study demonstrates the benefits of mixture-of-experts and FSA techniques to stabilize the generator's training process and promote the quality of generated samples. Besides, our models could generate sentences with high quality in terms of the semantic coherence and grammatical correctness of language, as per human evaluation. Furthermore, we empirically demonstrate that the proposed architecture overshadows most existing models in terms of quantitative and qualitative evaluation. 

To summarize, our main contributions are as follows:
\begin{itemize}
    \item We design a mixture-of-experts language GAN framework that integrates the multi-agent structure to enhance the expressive capacity of the generator network. 
    \item We utilize an auxiliary encoder to extract the latent embeddings and propose the Feature Statistics Alignment paradigm to endow the generator with fine-grained learning signals aside from the discriminator's feedback. We empirically demonstrate its effectiveness in promoting the quality of generated samples.
    \item Our method achieves new state-of-the-art results on three different benchmarks of language GANs, including synthetic data, MS COCO Image Captions dataset, and EMNLP2017 WMT News dataset. 
\end{itemize}


\section{Language GANs}
\label{sec:seqgan}
Adversarial sequence generation has attracted broad attention for its properties to solve the exposure bias issue suffered from maximum likelihood estimation (MLE) for generating language sequences. Based on the game theory, its goal is to train a generator network $G(z;\theta^{(G)})$ that produces samples from the data distribution $p_\textrm{data}(x)$ by decoding the randomly initialized starting token $z$ into the sequence $x=G(z;\theta^{(G)})$, where the training signal is provided by the discriminator network $D(x; \phi^{(D)})$ that is trained to distinguish between the samples drawn from the real data distribution $p_\textrm{data}$ and those produced by the generator. The minimax objective of adversarial training is formulated as:

\begin{align}
    \min_{\theta^{(G)}} \max_{\phi^{(D)}} {}&\mathbb{E}_{x \sim p_\textrm{data}} \big[\log D_{\phi}^{(D)} (x; \phi^{(D)})\big] \nonumber \\+&\mathbb{E}_{z \sim p_{z}} \big[\log \big(1 - D_{\phi}^{(D)}(G(z; \theta^{(G)})) \big) \big]
\end{align}

Despite the impressive results of GANs in the sequence generation~\cite{Yu2017SeqGANSG,Gulrajani2017ImprovedTO,Scialom2020ColdGANsTL}, there are still several fundamental issues in the GAN training: (a) Training instability, which arises from the intrinsic nature of minimax games in GANs; (b) Mode dropping, which is the fact that GANs only generate samples with limited patterns in the real data distribution instead of attending to diverse patterns~\cite{Chen2018AdversarialTG}; (c) Reward sparsity, which is because that it is easier to train the discriminator than the generator, making it difficult to acquire the instructive feedback~\cite{Zhou2020SelfAdversarialLW}.

Due to the non-differentiability of gradients caused by sampling operations between the generator and discriminator for sequence generation, the majority of previous works have resorted to reinforcement learning (RL) heuristics with Monte Carlo search to collect the credits from the discriminator. The usage of RL may further deteriorate the instability of model training and exacerbate the reward sparsity problem.
Gumbel-Softmax relaxation has proven to be an alternative to RL techniques~\cite{Kusner2016GANSFS,Nie2019RelGANRG}.
Therefore, we utilize the Gumbel-Softmax reparameterization instead of policy gradients to circumvent the unstable training of RL in our framework.

\section{Methodology}
\label{sec:me}
 Unlike conventional language GANs that only update the generator with only one type of learning signals, the proposed model collects extra update feedback by judging the distortion between real and generated data distributions, besides true-or-false comparative rewards. In this work, we design a language GAN framework, in which the generator is guided by two learning signals: comparative credits from the classifier, and distortion credits propagated from an auxiliary encoder. The former gauges the relativistic confidence of real samples compared with generated ones, while the latter measures the difference between latent feature statistics of real and generated samples.

\begin{figure*}[thb]
\begin{center}
\includegraphics[width=\textwidth]{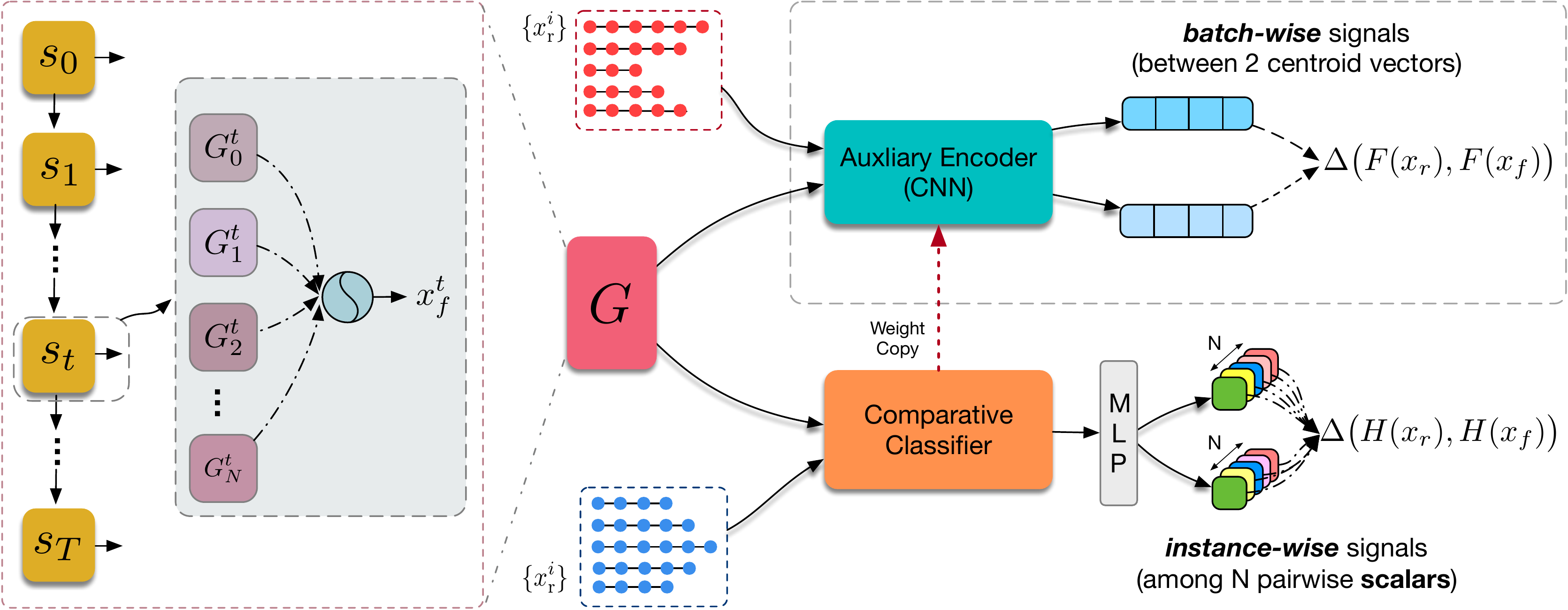}
\caption{Schematic illustration of Mixture-of-Experts language GANs.}
\label{fig:model}
\end{center}
\end{figure*}

As illustrated in Figure~\ref{fig:model}, MoEGAN consists of three components: (a) A generator that leverages multiple agents to collaboratively produce sequences to fool the discriminative classifier; (b) An auxiliary encoder that extracts latent embeddings from real and fake samples and thus renders fine-grained learning signals for generator updates; (c) A comparative classifier that measures the relative likelihood that given real samples are more authentic than fake data.

\subsection{Mixture-of-Experts Generator}
\label{subsec:ma-g}
We utilize multiple experts $\{ G_i \vert i \in (0, N_g) \}$ as generators, in which each expert receives as input previous generated tokens $x_{t-1}$ and produces the representation $r_i^t = G_i^t(x_{t-1})$ at the $t$-th time step. At the beginning, \emph{i.e.}, $t=1$, the input $x_0$ is the starting token $z$, which is a randomly initialized word embedding. The expert $G_i$ can be a kind of recurrent neural networks (RNNs), such as Long Short-Term Memory (LSTM) and Gated Recurrent Unit (GRU), which can autoregressively produce the representation of current tokens. 

Formally, the model distribution of $i$-th expert $G_i$ parameterized by $\theta_i$ can be defined as $q(r^t \vert x_i^{<t}; \theta_i)$. The overall distribution for the mixture-of-experts generator can be processed in various ways, such as concatenation or gating functions. Different experts jointly model the target hidden representations in collaboration to further generate realistic tokens. Since our work aims to identify the impact of the mixture-of-experts mechanism in sequence GANs, we simply take the expectation of them for aggregation and leave the exploration of various interaction ways in future work.

The model distribution of mutliple experts is formulated as:
\begin{equation}
    p(r^t \vert x^{<t}; \theta_t) = \frac{1}{N_g} \sum_{i=1}^{N_g} q(r_i^t \vert x_i^{<t}; \theta_i)
\end{equation}{where $\theta_t$ denotes the trainable parameters of the generator. This can also be treated as an ensemble of multiple agents, which is similar to~\cite{liu2018distilling}. In practice, each agent equivalently serves as a constituent to vote for candidate choices at every time step.}

Denoting the aggregated output of multiple agents as $Y^t \in \mathbb{R}^{D_g}$, the output probabilities over the $|V|$-dimensional output vocabulary at $t$-th time step is: 
\begin{equation}
    \pi^t = \textrm{softmax}(W_G Y^t)
\end{equation}{where $\pi^t = \{ \pi^t_1, \pi^t_2, \cdots, \pi^t_{|V|}\}$ represents the output probabilities of the vocabulary tokens, $W_G \in \mathbb{R}^{|V| \times D}$ denotes the trainable parameter.}

GANs have fallen short of discrete data generation primarily resulting from the incapability of gradient propagation passing from the discriminator to the generator, which is incurred by the non-differential sampling or argmax operations in between. Following~\cite{Kusner2016GANSFS,Nie2019RelGANRG}, we leverage the Gumbel-Softmax distribution~\cite{Jang2017CategoricalRW} by smoothly annealing to approximate the categorical distribution. The Gumbel-Max trick~\cite{maddison2014sampling} can be parameterized as:
\begin{equation}
    \mathbf{y^t} =  \textrm{one\_hot}\big( \arg\max_i [g_i + \log \pi^t_i] \big) \label{eq:gumbel-max}
\end{equation}{where $\{ g_i \vert i=1,\cdots,|V| \}$ are i.i.d from the Gumbel(0,1) distribution, that is, $g_i = - \log (- \log u_i)$ where $u_i$ is drawn from a standard uniform distribution Uniform(0,1). $\mathbf{y^t}$ represents the $|V|$-dimensional one hot encoding. }

The Gumbel-Softmax trick approximates the non-differential $\arg\max$ operation with the softmax function:
\begin{equation}
    \hat{y}^t_i = \frac{\exp\big( (\log(\pi^t_i) + g_i )/ \tau \big)  }{ \sum_{j=1}^{|V|} \exp\big( ( \log(\pi^t_j) + g_j )/ \tau \big) }
\end{equation}{where $\tau$ denotes the softmax temperature to modulate the exploitation and exploration during training. When $\tau$ approaches too high, the approximation is nearly equiprobable, encouraging the generator to explore different options. In contrast, the lower $\tau$ could discourage the exploration and tend to exploit during training. In particular, when $\tau \rightarrow 0$, $\hat{y}^t$ approaches the result of one-hot operator as in Eq.~\ref{eq:gumbel-max}, whereas $\hat{y}^t$ will degenerate into a uniform distribution when $\tau \rightarrow \infty$.}

\subsection{Auxiliary Encoder}
\label{subsec:aux-dec}

Given the input sequence of real samples (denoted as $x_r$) and that of generated ones (denoted as $x_f$), we use an auxiliary encoder $F$ to extract their latent feature embeddings and measure the difference of their respective latent distributions, \emph{i.e.}, $\Delta \big(F(x_r), F(x_f) \big)$. In practice, we adopt Convolutional Neural Networks (CNNs) as the auxiliary encoder.

Intuitively, aligning the statistics of embedded feature representations increases the model capacity to capture various modes of the data distribution. For brevity, we utilize the first-order mean statistics in our framework and leave the higher-order statistics for future work.

We propose Feature Statistics Alignment (FSA), which measures the Euclidean distance between the minibatch centroids of real and fake feature representations. Given mini-batches of real data samples with the batch size of $N_f$, we formulate the FSA distance as:
\begin{align}
        &{}\Delta \big(F(x_r), F(x_f) \big) = \nonumber \\ &{}\big\Vert \mathbb{E}_{x_r \sim p_\textrm{data}} \big[ F(x_r) \big] - \mathbb{E}_{z \sim p_{z}} \big[ F(G(z; \theta^{(G)}))\big] \big\Vert_2  \label{eq:fsa}
\end{align}


By forcing the mean statistics of fake data to be close to the real samples, the generator could receive more informative signals during the training process. 

Introducing an extra auxiliary encoder could cost additional computing resources and may slow down the overall training process. To mitigate this issue, we do not update the weight of the proposed encoder during training. Instead, we copy the parameters from the comparative classifier after each training iteration and keep them fixed to extract the latent embeddings (This requires identical model structures between the auxiliary encoder and the comparative classifier).



\subsection{Comparative Classifier}
\label{subsec:disc}
Following the relativistic discriminator~\cite{JolicoeurMartineau2019TheRD}, we take into account the relativistic confidence of given real samples using a comparative classifier. 

Given the discriminator $H$, the mismatch between real and fake inputs measures the relative confidence that given real data are more realistic than randomly sampled fake data , which can be formulated as:
\begin{equation}
    \Delta \big( H(x_r), H(x_f) \big) = H(x_r) - H(G(z; \theta^{(G)}))
\end{equation}

The comparative discriminator aims to maximize the likelihood that given real instances are more authentic than generated ones, whereas the generator runs counter to it. The loss function of the comparative discriminator is defined as:
\begin{equation}
    \mathcal{L}_\textrm{D} = - \mathbb{E}_{x \sim p_{\textrm{data}}, z \sim p_z} \log a [\Delta \big( H(x_r), H(x_f) \big)] \label{eq:loss-d} 
\end{equation}{where $a$ represents the activation function to be relativistic (we use sigmoid function in our experiments). }

\subsection{Training}
\label{subsec:train}
\paragraph{Generator's Training Objectives.} In language GANs, the generator is more difficult to train than the discriminator, resulting in training instability and reward sparsity. To relieve these issues, we endow the optimization of the generator with signals from the auxiliary encoder. The loss function is thus:
\begin{equation}
    \mathcal{L}_\textrm{G} = - \mathcal{L}_\textrm{D} + \Delta \big(F(x_r), F(x_f) \big) \label{eq:loss-g} 
\end{equation}{The goal of the discriminator is to maximize the gap between the generated and real data (Eq.~\ref{eq:loss-d}), whereas the generator jointly considers two different aspects simultaneously: it not only competes with the discriminator by maximizing the gap in terms of relativistic signals but takes into account feature distortion signals (Eq.~\ref{eq:loss-g}). This could pass more instructive feedback only to the generator, and also prevent the discriminator from being overtrained.}

\paragraph{Adversarial Training Algorithm.} Algorithm~\ref{alg:gan} illustrates the overall training process of the proposed framework. The Relativistic Discriminator and the generator could reach the Nash Equilibrium when the generator could fool the discriminator into accepting its output as being true. Since the discriminator network is easy to be overtrained, we do not pretrain it but only pretrain the generator using MLE for few epochs.

\begin{algorithm}[ht]
\begin{algorithmic}[1]
  \STATE {\bfseries Require:} generator $G_\theta$; discriminator $D_{\phi}$; auxiliary encoder $F_\omega$; samples of real data $\mathbb{S}$; generator training step $g$; discriminator training step $k$; the generator pretraining epochs $m$.
  \STATE Pretrain $G_\theta$ using MLE on $\mathbb{S}$ for $m$ epochs
    \REPEAT
    \STATE Copy parameters from $D_{\phi}$ to $F_\omega$
    \FOR{$g$ steps}
    \STATE Sample a minibatch from real data $\mathbb{S}$
    \STATE Generate a minibatch of samples $x_f \sim G_\theta$
    \STATE Compute $\Delta \big(F_\omega(x_r), F_\omega(x_f) \big)$ with Eq.(\ref{eq:fsa})
    \STATE Update $G_\theta$ via Eq.(\ref{eq:loss-g})
    \ENDFOR
    \FOR{$k$ steps}
    \STATE Sample a minibatch from real data $\mathbb{S}$
    \STATE Sample a minibatch from the generated data
    \STATE Train the discriminator $D_{\phi}$ by Eq.(\ref{eq:loss-d})
    \ENDFOR
    \UNTIL{convergence}
\end{algorithmic} 
  \caption{Adversarial Training of MoEGAN}
  \label{alg:gan} 
\end{algorithm}

\section{Experiments}
\label{sec:exp}
\subsection{Experimental Settings}

\paragraph{Dataset.} Following \cite{Lin2017AdversarialRF,Guo2018LongTG,Nie2019RelGANRG,Zhou2020SelfAdversarialLW}, we evaluate the proposed framework based on the Texygen benchmark platform~\cite{Zhu2018TexygenAB} for adversarial text generation. Experiments were conducted on synthetic and real datasets: (a) synthetic data, which is generated by an oracle single-layer LSTM as in ~\cite{Yu2017SeqGANSG}; (b) MS COCO Image Caption dataset~\cite{chen2015microsoft}; (c) EMNLP WMT 2017 News dataset~\cite{Guo2018LongTG}. Table~\ref{tab:dataset} summarizes the statistics of benchmark datasets for evaluation.

\begin{table}[thb]
\centering
\resizebox{\linewidth}{!}{%
\begin{tabular}{@{}lrrrr@{}}
\toprule
dataset & vocabulary size & sequence length & training set & test set \\ \midrule
synthetic data & 5,000 & 20 / 40 & 10,000 & 10,000 \\
MS COCO & 4,657 & 37 & 10,000 & 10,000 \\
EMNLP2017 WMT News & 5,255 & 51 & 278,586 & 10,000 \\ \bottomrule
\end{tabular}%
}
\caption{Summary of experimental datasets.}
\label{tab:dataset}
\end{table}

\paragraph{Evaluation Metrics.} (1) For synthetic data experiments, we utilize a single-layer LSTM initialized by standard normal distribution as the oracle model, which is used to generate 10,000 samples of length 20 and 40 respectively as real data. We use the negative log-likelihood (NLL) under the oracle data distribution for evaluation, termed NLL$_\textrm{oracle}$. (2) For real data, BLEU scores
~\cite{papineni2002bleu} are used to evaluate the n-gram statistics overlapping on the whole dataset.  (3) To measure the diversity of generated samples, the NLL of the generator (denoted as NLL$_\textrm{gen}$) is used by computing the NLL of reference samples in the test set by the generator. (4) Considering that BLEU scores always focus on the local text statistics and may be insufficient for evaluating the overall quality of texts, we conducted additional human evaluation via crowdsourcing on comparison models.

\paragraph{Baselines.} We adopt MLE and other state-of-the-art models as baselines, involving SeqGAN~\cite{Yu2017SeqGANSG}, RankGAN~\cite{Lin2017AdversarialRF}, LeakGAN~\cite{Guo2018LongTG}, RelGAN~\cite{Nie2019RelGANRG}, Self-Adversarial Learning (SAL)~\cite{Zhou2020SelfAdversarialLW}, and Counter-Contrastive Learning GAN (CCL)~\cite{chai-etal-2021-counter-contrastive}.

\paragraph{Implementation Details.} We adopt the Relational Memory Core (RMC)~\cite{santoro2018relational} as the agent architecture of the generator. As for the discriminator and auxiliary encoder, we use CNN architecture~\cite{kim2014convolutional} for input sequences. See Appendix~\ref{ap:exp_details} for more details.

\subsection{Experimental Results}
\label{sec:res}
\paragraph{Synthetic Data.}
Table~\ref{tab:synthetic-data} illustrates the performance of different models on NLL$_\textrm{oracle}$. Our models outperform other baseline models in terms of the generated sample quality, demonstrating the effectiveness of our proposed method. We empirically found that our models achieve the superior performance with the sequence length of 20 / 40 when the number of agent $N_g$ takes 2 and 3 respectively. As to the diversity, our method outperforms or achieves competitive NLL$_\textrm{gen}$ score compared with baselines (see Appendix~\ref{ap:synthetic} for details).

\begin{table*}[ht]
\centering
\resizebox{\linewidth}{!}{%
\begin{tabular}{@{}c|rrrrrrr| c |r@{}}
\toprule
Length & MLE & SeqGAN & RankGAN & LeakGAN & RelGAN & SAL & CCL & Ours  & Real \\ \midrule
20 & 9.038 & 8.736 & 8.247 & 7.038 & 6.680{\small$\pm$0.343} & 7.71{\small$\pm$0.17} & 6.77{\small$\pm$0.34} & \textbf{5.835}{\small$\pm$0.353} & 5.750 \\
40 & 10.411 & 10.310 & 9.958 & 7.191 & 6.765{\small$\pm$0.026} &  9.31{\small$\pm$0.03} & 6.65{\small$\pm$0.14} & \textbf{5.713}{\small$\pm$0.289}  & 4.071 \\ \bottomrule
\end{tabular}%
} 
\caption{The NLL\textsubscript{oracle} performance of different models on the synthetic dataset with the sequence length of 20 and 40 respectively, where $\tau=1$ for all model settings, $N_g=2 / 3$ for models with the sequence length of 20 and 40. For NLL, the lower, the better. All results of baseline models are obtained from original papers.}
\label{tab:synthetic-data} 
\end{table*}

\begin{table*}[]
\resizebox{\textwidth}{!}{%
\begin{tabular}{@{}l|llll|l|lllll@{}}
\toprule
\multirow{2}{*}{Model} & \multicolumn{5}{c|}{MS COCO Image Captions}                                                                                                                                  & \multicolumn{5}{c}{EMNLP2017 WMT News}                                                                                                                                                   \\ \cmidrule(l){2-11} 
                       & BLEU-2                           & BLEU-3                           & BLEU-4                           & BLEU-5                           & NLL\textsubscript{gen}           & BLEU-2                           & BLEU-3                           & BLEU-4                           & \multicolumn{1}{l|}{BLEU-5}                           & NLL\textsubscript{gen}  \\ \midrule
MLE                    & 0.731                            & 0.497                            & 0.305                            & 0.189                            & 0.718                            & 0.768                            & 0.473                            & 0.240                            & \multicolumn{1}{l|}{0.126}                            & 2.382                   \\
SeqGAN                 & 0.745                            & 0.498                            & 0.294                            & 0.180                            & 1.082                            & 0.777                            & 0.491                            & 0.261                            & \multicolumn{1}{l|}{0.138}                            & 2.773                   \\
RankGAN                & 0.743                            & 0.467                            & 0.264                            & 0.156                            & 1.344                            & 0.727                            & 0.435                            & 0.209                            & \multicolumn{1}{l|}{0.101}                            & 3.345                   \\
LeakGAN                & 0.746                            & 0.528                            & 0.355                            & 0.230                            & 0.679                            & 0.826                            & 0.645                            & 0.437                            & \multicolumn{1}{l|}{0.272}                            & \textbf{2.356}          \\
RelGAN            & 0.849{\small$\pm$0.030}          & 0.687{\small$\pm$0.047}          & 0.502{\small$\pm$0.048}          & 0.331{\small$\pm$0.044}          & 0.756{\small$\pm$0.054}          & 0.881{\small$\pm$0.013}          & 0.705{\small$\pm$0.019}          & 0.501{\small$\pm$0.023}          & \multicolumn{1}{l|}{0.319{\small$\pm$0.018}}          & 2.482{\small$\pm$0.031} \\
SAL                   & 0.785{\small$\pm$0.02}           & 0.581{\small$\pm$0.03}           & 0.362{\small$\pm$0.02}           & 0.227{\small$\pm$0.02}           & 0.873{\small$\pm$0.02}           & 0.788{\small$\pm$0.02}           & 0.523{\small$\pm$0.02}           & 0.281{\small$\pm$0.02}           & \multicolumn{1}{l|}{0.149{\small$\pm$0.02}}           & 2.578{\small$\pm$0.04}\\
CCL               & 0.871{\small$\pm$0.032} & 0.715{\small$\pm$0.050} & 0.538{\small$\pm$0.068} & 0.399{\small$\pm$0.082} & 0.630{\small$\pm$0.103}   & 0.903 {\small$\pm$0.016} & 0.749{\small$\pm$0.022} & 0.525{\small$\pm$0.017} & \multicolumn{1}{l|}{0.324{\small$\pm$0.008}} & 2.818{\small$\pm$0.499} \\ 
\midrule
Ours                   & \textbf{0.963}{\small$\pm$0.020} & \textbf{0.902}{\small$\pm$0.059} & \textbf{0.814}{\small$\pm$0.072} & \textbf{0.695}{\small$\pm$0.076} & \textbf{0.639}{\small$\pm$0.027} & \textbf{0.910}{\small$\pm$0.015} & \textbf{0.769}{\small$\pm$0.021} & \textbf{0.568}{\small$\pm$0.026} & \multicolumn{1}{l|}{\textbf{0.374}{\small$\pm$0.023}} & 2.480{\small$\pm$0.025} \\ \bottomrule
\end{tabular}%
}
\caption{The BLEU and NLL\textsubscript{gen} performance on real data, in which we adopt $\tau=0.01$ and $N_g=2$. For all BLEU scores, the higher, the better. For NLL, the lower, the better.}
\label{tab:real-data}
\end{table*}

\paragraph{Real Data.} To further verify the performance on real data, we run and evaluate our model on MS COCO image caption and EMNLP2017 WMT News dataset. The data preprocessing remains the same as Texygen~\cite{Zhu2018TexygenAB}. Table~\ref{tab:real-data} compares the results of our model with baselines in terms of both the quality (BLEU) and diversity (NLL\textsubscript{gen}) metrics. Overall, our model exceeds or matches the comparison models on automatic evaluation metrics.

For \textbf{quality metrics}, SeqGAN and RankGAN outperform the MLE on BLEU scores with short n-gram spans, such as BLEU-2, but are inferior to MLE when comparing long-term spans in metrics such as BLEU-4/5, which may result from the lack of informative signals for the generator update. Nevertheless, LeakGAN, RelGAN, SAL, and CCL surpass the MLE method on all BLEU scores. This implies that both leakage information and comparative feedback can enhance the adversarial training. The improvements of our model over MLE are even larger than comparison models, demonstrating the benefits of exploiting the mixture-of-experts and FSA paradigm on language GANs. In terms of the \textbf{diversity metric}, most baselines behave not as well as MLE, except LeakGAN. This is because LeakGAN leverages the internal information from discriminators as the guide signal, assisting the original learning credits collected from the discriminator. Regarding the diversity metric, the proposed model slightly outranks LeakGAN and MLE on the MS COCO Image Caption dataset (maximum length 37), but achieves similar results on the EMNLP 2017 WMT News dataset (maximum length 51), since it is even difficult to produce the informative signals for long sequences. By leveraging the proposed methods, our model greatly outperforms these models and yields long sequences with promising qualities while generates samples with similar diversity.  It implies that our model is adept at generating high-quality sentences under the guidance of introduced latent-feature-enhanced signals.

\paragraph{Human Evaluation.} Apart from the automatic evaluation, we also conducted human evaluation on MS COCO dataset. We randomly sampled 100 sentences from each model, then asked ten different people to score them in terms of grammatical correctness and meaningfulness on a scale of 1-5 after anonymizing the model's identity. Our model received the highest score in comparison with other baselines as shown in Table~\ref{tab:human_scores}. Please see Appendix~\ref{ap:human_eval} for details.

\begin{table}[thb]
\centering
\resizebox{\columnwidth}{!}{%
\begin{tabular}{@{}c|cccc@{}}
\toprule
Model       & MLE               & SeqGAN            & RankGAN           & LeakGAN           \\ \midrule
Human score & 3.126 $\pm$ 0.388 & 3.056 $\pm$ 0.349 & 3.017 $\pm$ 0.386 & 3.005 $\pm$ 0.353 \\ \midrule\midrule
Model       & MaliGAN           & TextGAN           & RelGAN            & Ours              \\ \midrule
Human score & 3.014 $\pm$ 0.385 & 1.967 $\pm$ 0.421 & 3.709 $\pm$ 0.441 &   \textbf{4.077 $\pm$ 0.348}                \\ \bottomrule
\end{tabular}%
}
\caption{Mean and standard deviation results of different models by human evaluation on MS COCO Image Caption dataset.}
\label{tab:human_scores}
\end{table}

\begin{figure}[thb]
\vskip -3mm
\begin{center}
\includegraphics[width=\columnwidth]{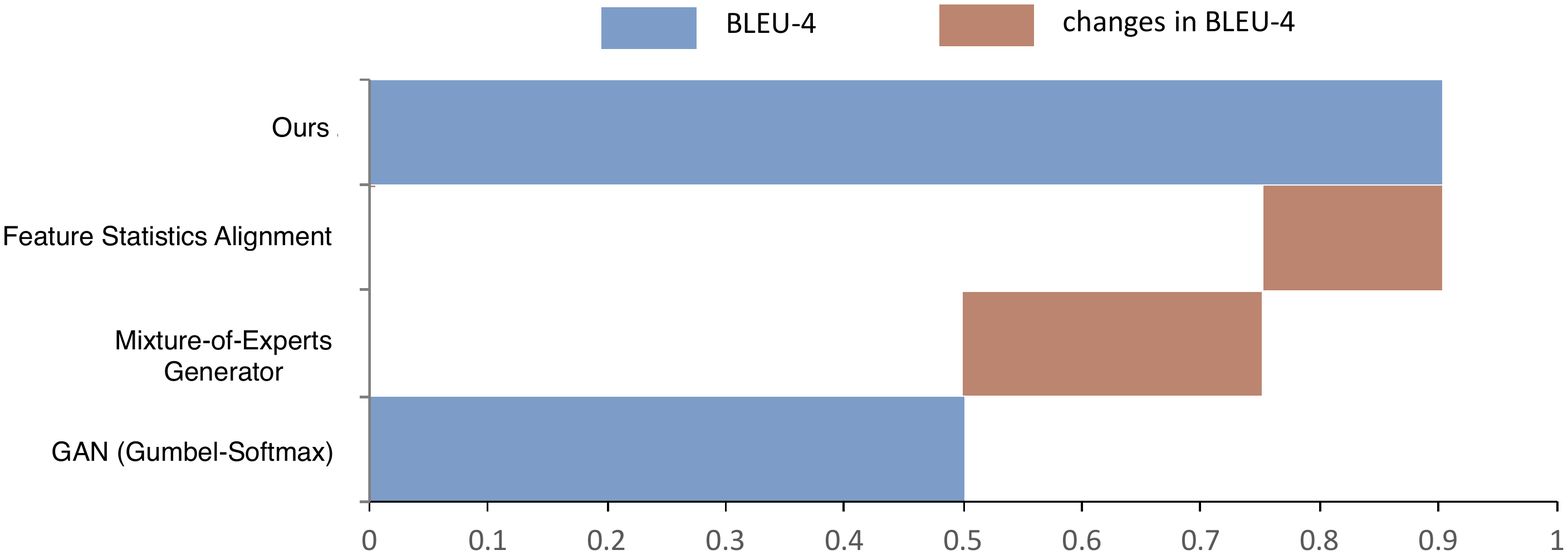}
\caption{Ablation study.}
\label{fig:rel_ablation}
\end{center}
\vskip -3mm
\end{figure}

\begin{figure}[thb]
     \begin{subfigure}[b]{0.48\columnwidth}
         \centering
         \includegraphics[width=\textwidth]{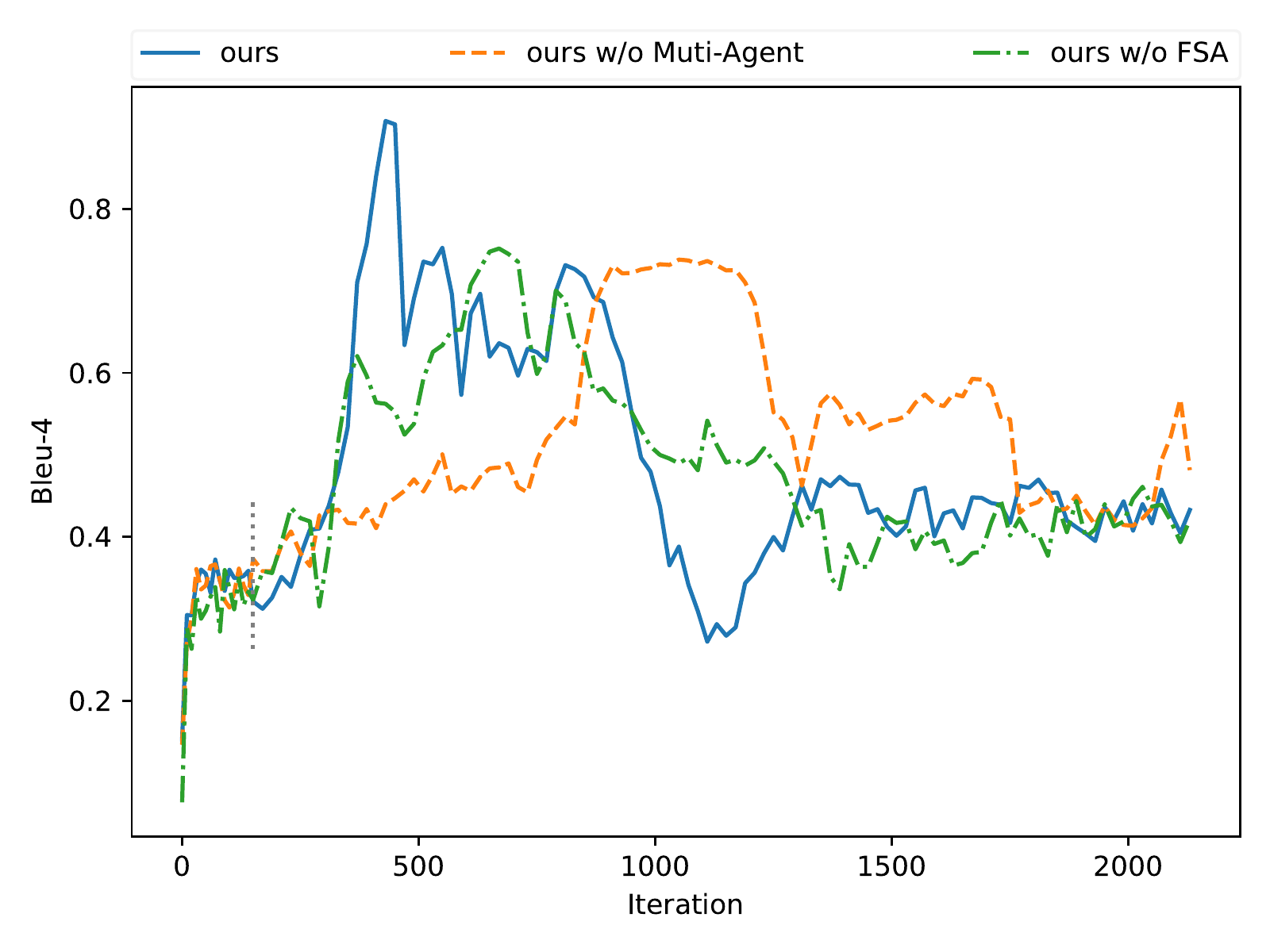}
     \end{subfigure}
     \hfill
     \begin{subfigure}[b]{0.48\columnwidth}
         \centering
         \includegraphics[width=\textwidth]{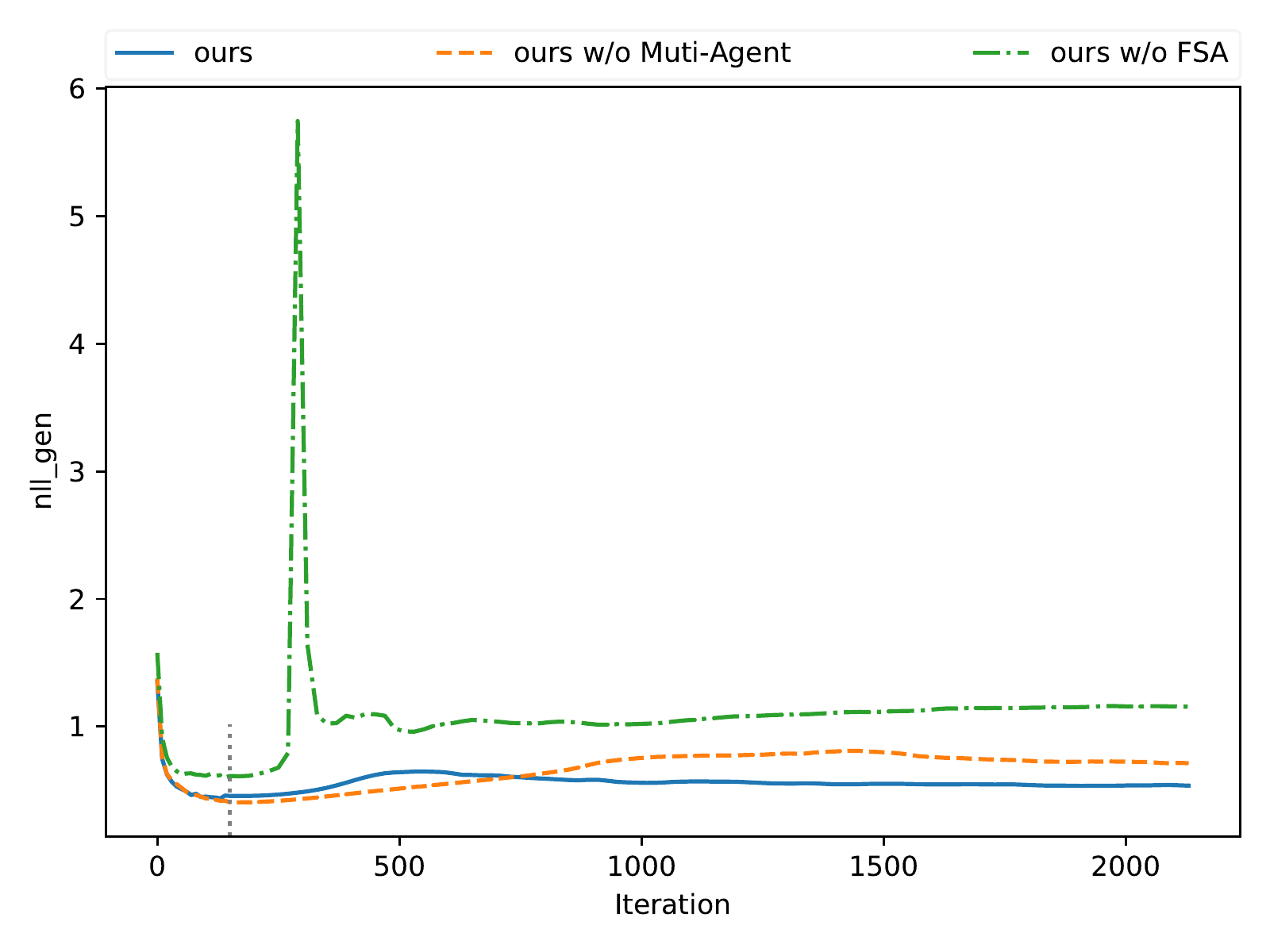}
     \end{subfigure} 
        \caption{The BLEU-4 (left) and NLL\textsubscript{gen} (right) performance of our models w/ and w/o  mixture-of-experts and FSA on MS COCO dataset. Vertical dash lines indicate the end of generator pretraining. 
        }
        \label{fig:COCO_ablation}
\end{figure}

\begin{figure*}[t]
    \centering
    \begin{subfigure}[b]{0.22\linewidth}
        \centering
        \includegraphics[width=\textwidth]{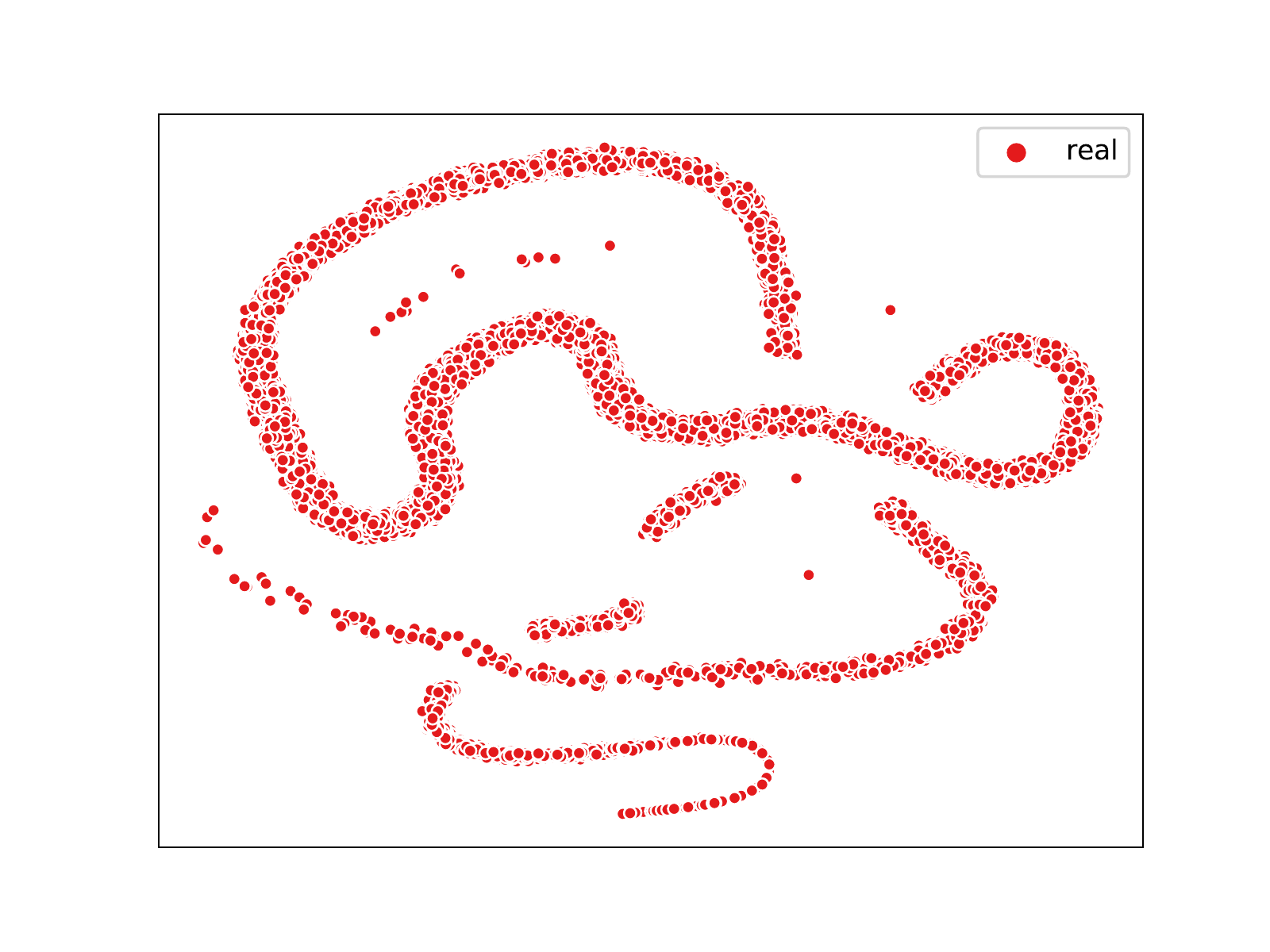}
        \caption[]%
        {{ Real data w/ FSA}}    
        \label{fig:real-ours}
    \end{subfigure}
    \hfill
    \begin{subfigure}[b]{0.22\linewidth}  
        \centering 
        \includegraphics[width=\textwidth]{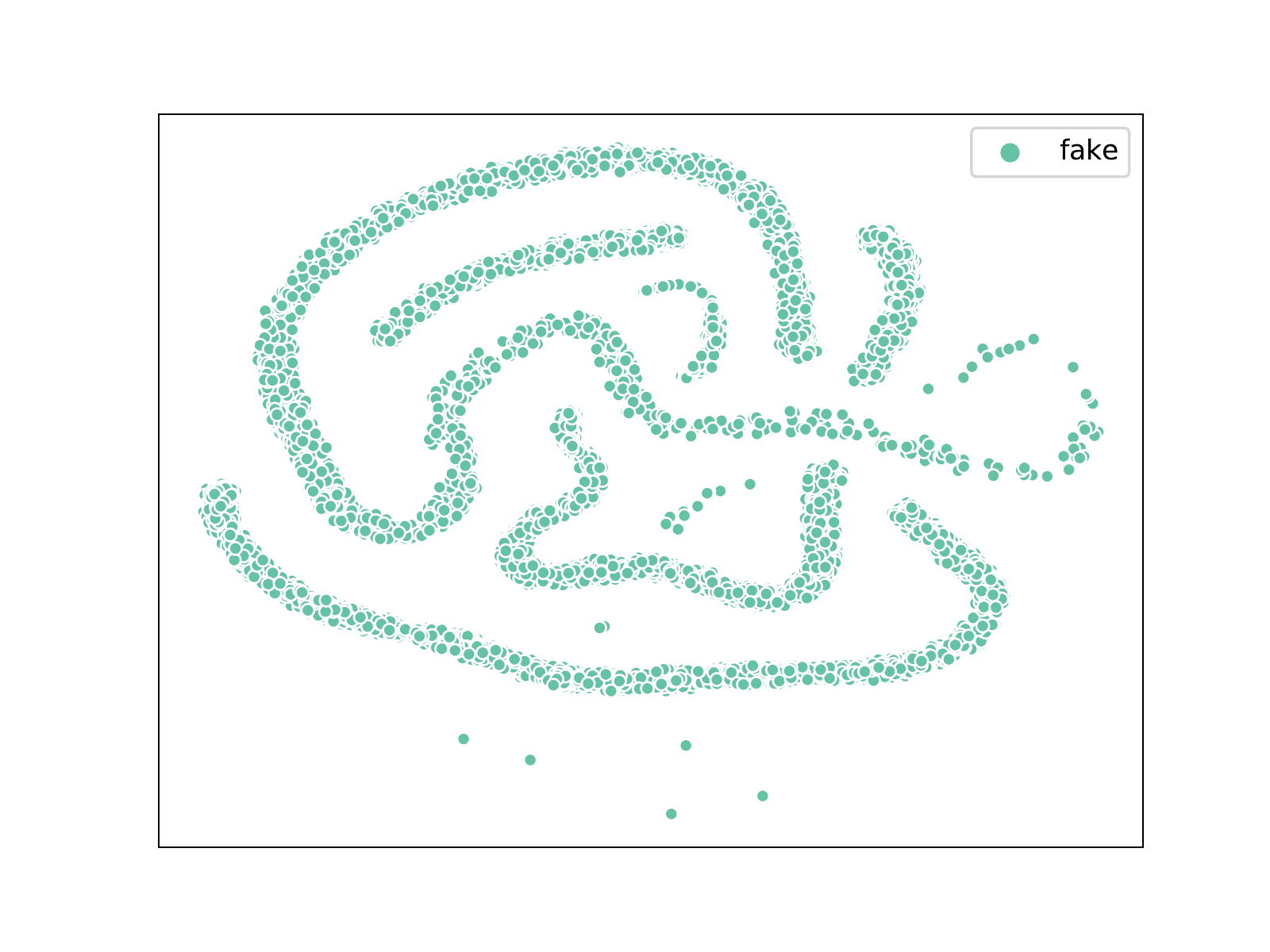}
        \caption[]%
        {{ Fake data w/ FSA}}    
        \label{fig:fake-ours}
    \end{subfigure}
    \hfill
    \begin{subfigure}[b]{0.22\linewidth}   
        \centering 
        \includegraphics[width=\textwidth]{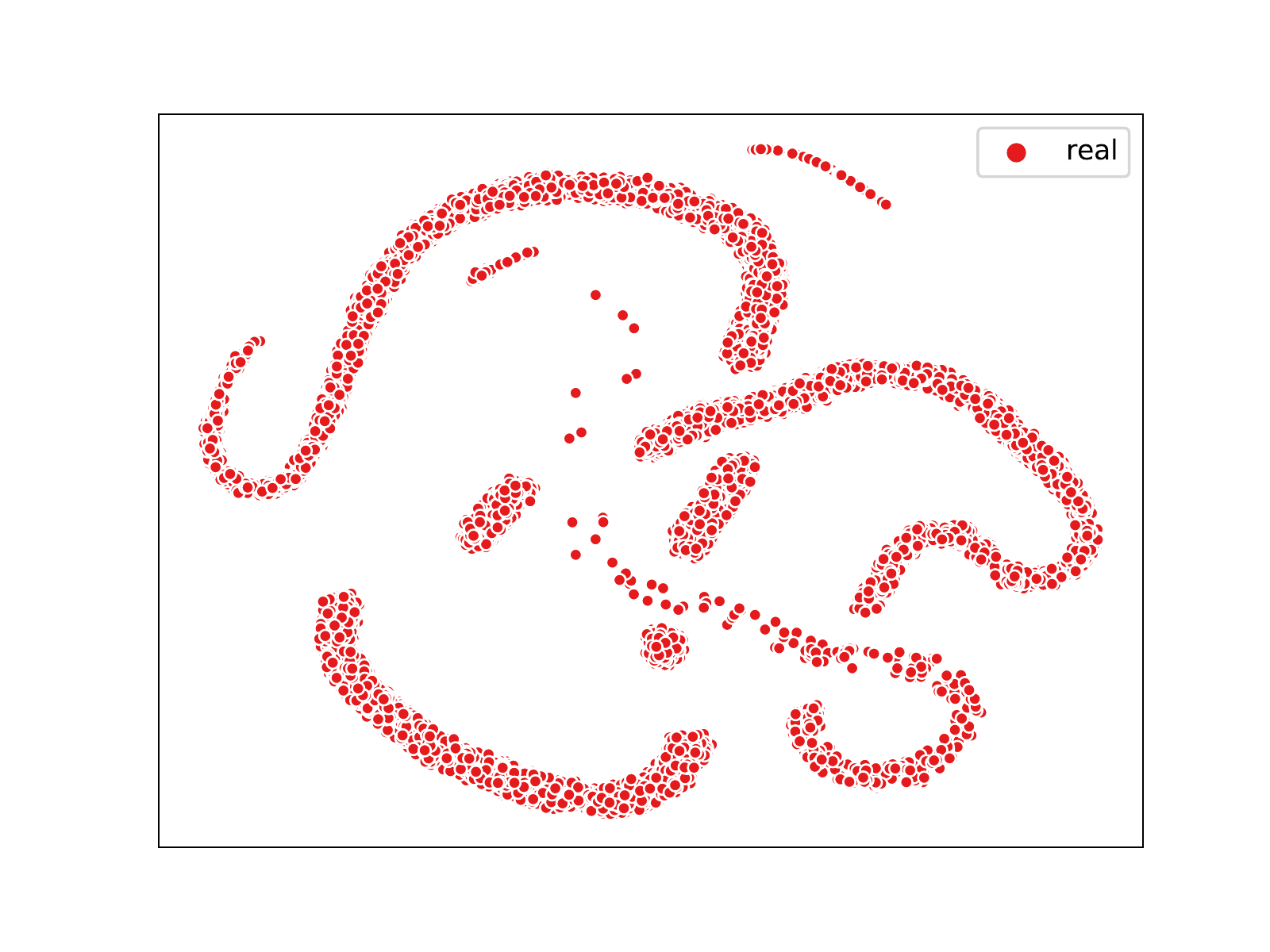}
        \caption[]%
        {{ Real data w/o FSA}}    
        \label{fig:real-comp}
    \end{subfigure}
    \hfill
    \begin{subfigure}[b]{0.22\linewidth}   
        \centering 
        \includegraphics[width=\textwidth]{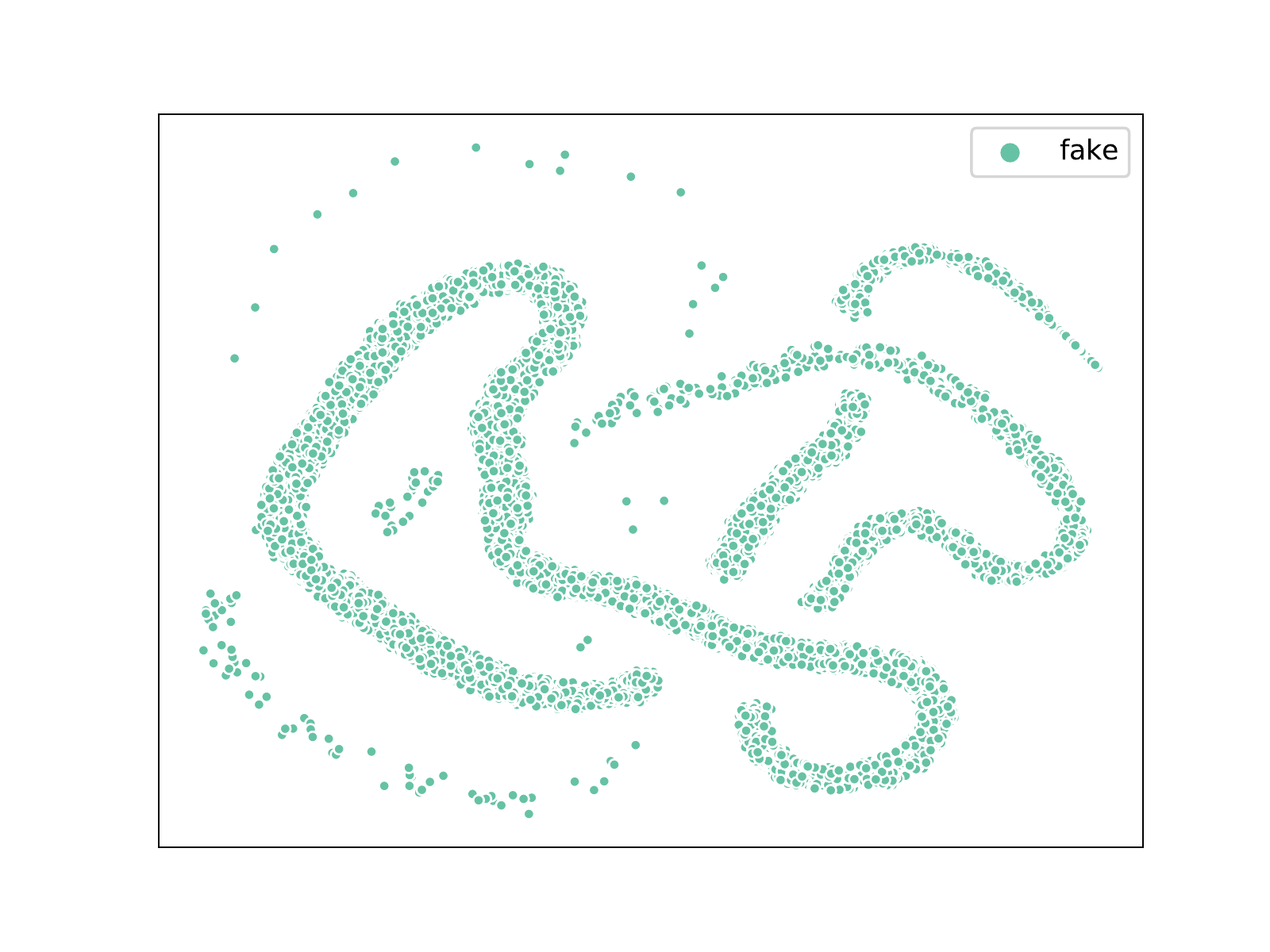}
        \caption[]%
        {{ Fake data w/o FSA}}    
        \label{fig:fake-comp}
    \end{subfigure}
    \caption[ ]
    { T-SNE plot of real and fake embeddings w/ and w/o FSA. (a)(b): w/ FSA; (c)(d): w/o FSA.
    } 
    \label{fig:TSNE_FSA}
\end{figure*}

\paragraph{Ablation Study.} To examine the benefits of proposed components in adversarial sentence generation, we conducted ablation tests by removing particular modules. Figure~\ref{fig:rel_ablation} demonstrates the relative importance of each component of our models with an ablation test. The usage of the mixture-of-experts generator results in the most significant performance gain, followed by the FSA. We empirically found that proposed approaches are able to boost the performance due to accurate estimates of feature statistics and stabilizing effects during the adversarial training. Note that following experiments are conducted on the MS COCO dataset if not otherwise specified.

Figure~\ref{fig:COCO_ablation} illustrates the ablation test of mixture-of-experts generator and FSA in terms of the BLEU-4 score (See Appendix~\ref{ap:ablation} for all results) and NLL\textsubscript{gen}. It can be observed that the ablation of either the mixture-of-experts or FSA approach deteriorates the model performance: the full model achieves a higher BLEU score and lower NLL\textsubscript{gen}. Note that without the mixture-of-experts generator, the increasing trend of BLEU scores tends to be slow, whereas FSA may contribute more to the sample diversity in contrast to mixture-of-experts paradigm. This is because the multiple experts are able to enrich the representation capacity of generators, while the FSA paradigm could provide the consecutive ``fined-grained'' smoother learning signals to update the generator.

\paragraph{Auxiliary Feature Visualization.} To visually verify the FSA's impact, we plot sampled latent features of real and fake data using t-SNE. Figure~\ref{fig:real-comp} and~\ref{fig:fake-comp} compares latent representations of real and generated data using our models without FSA, showing a relative mismatch between visualized feature embeddings, especially the missing area on the top left corner in Figure~\ref{fig:fake-comp}. Meanwhile, there are several surplus embedding segments in fake data (Figure~\ref{fig:fake-comp}) but not in real data (Figure~\ref{fig:real-comp}). On the contrary, Figure~\ref{fig:real-ours} and~\ref{fig:fake-ours} reveal a larger overlapping area between real and fake embeddings generated by our model with FSA, indicating the benefit of the FSA on matching the distribution of real and generated data. This can further support previous automatic evaluation results.



\begin{figure}[thb]
     \begin{subfigure}[b]{0.48\columnwidth}
         \centering
         \includegraphics[width=\textwidth]{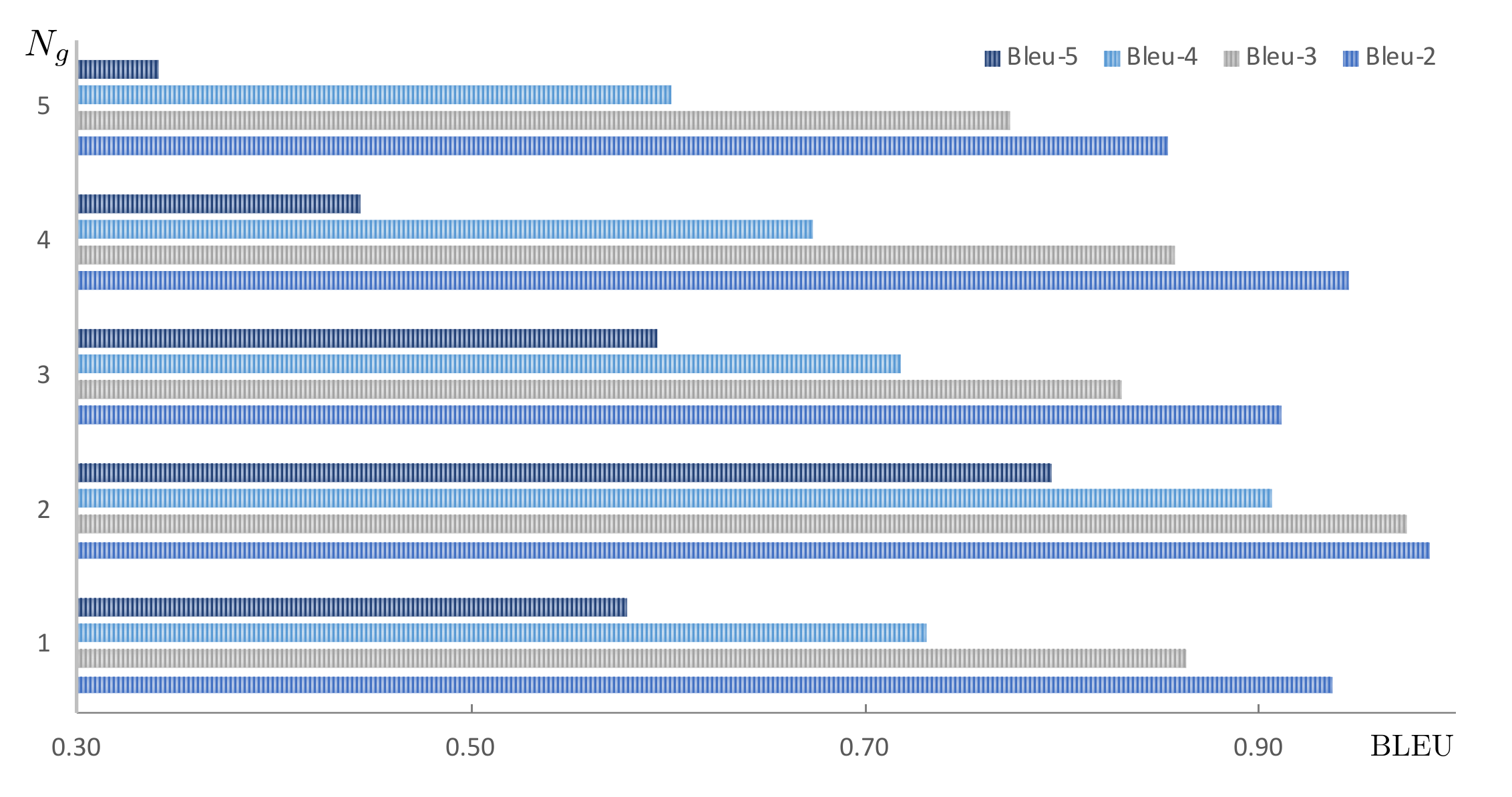}
         \caption{Impact of the expert number.}
         \label{fig:MA-impact}
     \end{subfigure}
     \hfill
     \begin{subfigure}[b]{0.48\columnwidth}
         \centering
         \includegraphics[width=\textwidth]{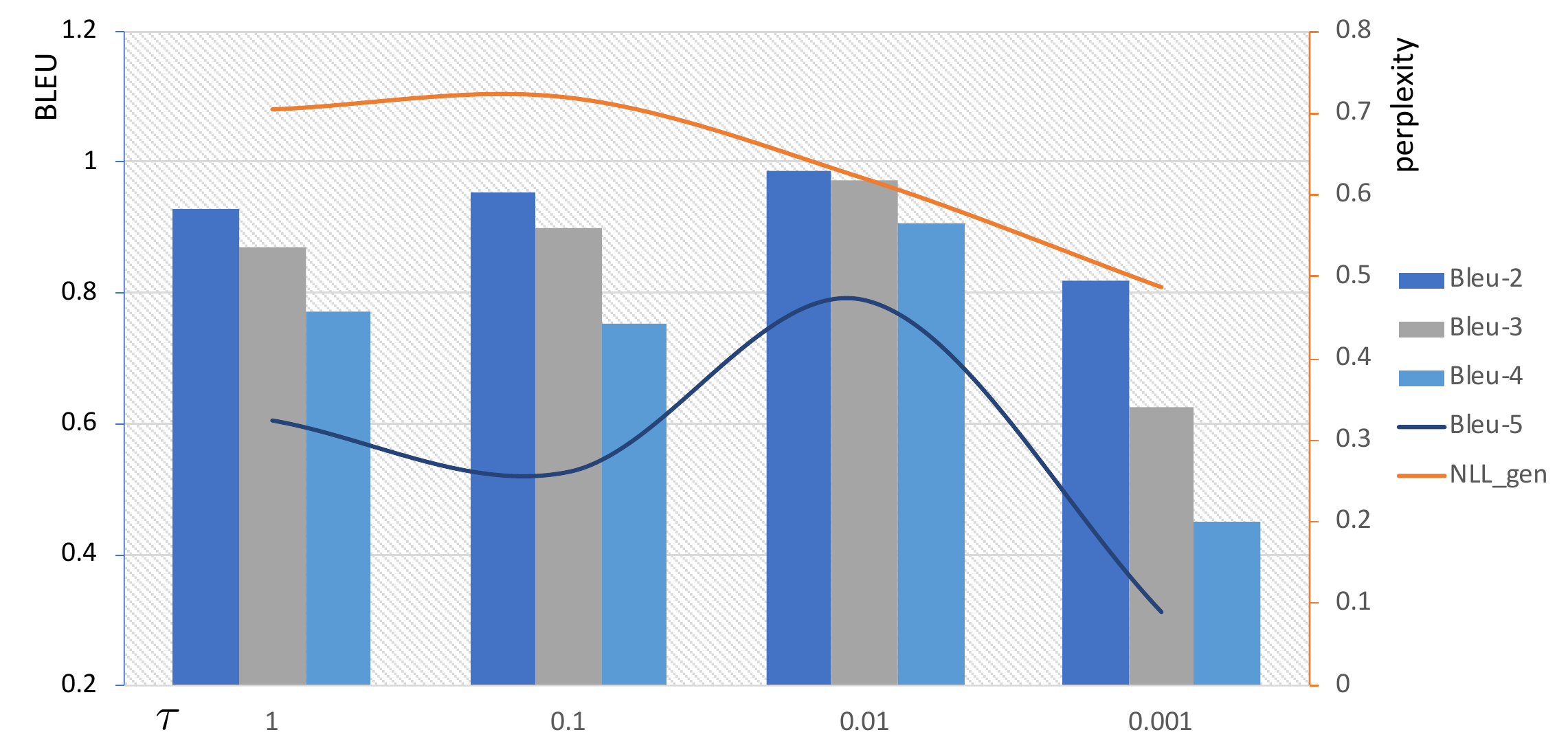}
         \caption{Impact of the temperature $\tau$.}
         \label{fig:temperature-impact}
     \end{subfigure} 
        \caption{The impact of agent number $N_g$ (a) and Gumbel-Softmax temperature $\tau$ (b) on model performance. }
        \label{fig:impact}
\end{figure}

\paragraph{Impact of Hyperparameters.} Figure~\ref{fig:MA-impact} displays the impact of agent number $N_g$ in the generator network on MS COCO Image Caption dataset. We can see that all BLEU scores reach their top when $N_g = 2$. We guess this is because we adopt an identical model structure for each generation agent, which may saturate with the increase of agent numbers. 
Figure~\ref{fig:temperature-impact} manifests the influence of the Gumbel-Softmax temperature $\tau$ on model performance. It is obvious that our models attain the best BLEU scores when $\tau=0.01$ whereas get lower NLL\textsubscript{gen} when $\tau=0.001$. To take the trade-off between the quality and diversity, we adopt $\tau=0.01$ in our optimal hyperparameter settings.


\section{Related Work}
\label{sec:bg}
There has been a large category of GANs for sequence generation, which heavily rely on the RL paradigm. SeqGAN~\cite{Yu2017SeqGANSG} regards the sequence generation as a Markov decision making process, estimates rewards via Monte Carlo search, and trains the generator with policy gradient. RankGAN~\cite{Lin2017AdversarialRF} and SAL~\cite{Zhou2020SelfAdversarialLW} replace the binary classifier in the discriminator as comparative discriminators to focus on relations between constructed pairs. MaliGAN~\cite{Che2017MaximumLikelihoodAD} utilizes the information in the discriminator as an additional source of training signals in the MLE objective to reduce the variance of gradients. LeakGAN~\cite{Guo2018LongTG} leaks the intermediate information via a manager to guide the generator, which is inspired by hierarchical RL. ColdGAN~\cite{Scialom2020ColdGANsTL} integrates the advance of importance sampling, RL algorithm to finetune pretrained models.

Another approach uses non-RL methods for adversarial sequence generation by either approximating the categorical sampling or directly using the continuous latent representation. TextGAN~\cite{Zhang2017AdversarialFM} uses feature matching via a kernelized discrepancy in the Reproducing Kernel Hilbert Space. FMGAN~\cite{Chen2018AdversarialTG} proposes to match feature distributions using a Feature-Mover's Distance. Both apply annealed soft-argmax for approximation. ARAML~\cite{ke2019araml} utilizes Reward Augmented Maximum Likelihood by sampling from the stationary distribution to acquire rewards. 
Gumbel-Softmax (GS) GAN~\cite{Kusner2016GANSFS} and RelGAN~\cite{Nie2019RelGANRG} prove the usefulness of Gumbel-Softmax in language GANs. \citet{chai-etal-2021-counter-contrastive} propose the counter-contrastive learning objective to learn contrastive signals by explicitly comparing real and fake samples.  
However, improving the training of language GANs still remains an open problem. Our model aims to promote GS-GAN with the proposed techniques to boost the training of language GANs. We also report a list of techniques we tried but proved to be unsuccessful or unnecessary in Appendix~\ref{ap:neg_results}.   To the best of our knowledge, the proposed model is the first that employs mixture-of-experts techniques in language GANs.

\section{Conclusion}
\label{sec:concl}
We propose an adversarial training framework for discrete sequence generation, by leveraging the advance of mixture-of-experts generator and Feature Statistics Alignment. Our model empirically shows superior performance in terms of quantitative and human evaluation. In the future, it is promising to extend our method to large language models.




\bibliography{anthology,custom}
\bibliographystyle{acl_natbib}

\appendix

\section{Implementation Details}
\label{ap:exp_details}
\paragraph{Mixture-of-Experts Generator.}
For each agent in the generator, we adopt the Relational Memory Core (RMC)~\cite{santoro2018relational}, setting the memory size as 256, the memory slot number as 1, the attention head number as 2. The input embedding dimension is set to 32. Similar to~\cite{hoang2018mgan}, we utilize parameter sharing among different agents to reduce the computing budget.

\paragraph{Comparative Classifier \& Auxiliary Encoder.} 
 For the Comparative Classifier and Auxiliary Encoder, we employ the multi-channel convolution using multiple filters with various window sizes to extract the distinct n-gram features, followed by a max-over-time pooling operation to gather the most salient features, \emph{i.e.}, features with the highest value for each feature map.
The input embedding dimension for the discriminator is set to 64. We adopt the filter size of $\{2,3,4,5\}$ with the number of 300 channels for each. A max-over-time pooling is adopted after the convolution layer. Afterward, a highway layer that is identical to SeqGAN~\cite{Yu2017SeqGANSG} is used followed by a linear transformation with the dimension of 100. Finally, apply a linear transformation to get the final logits. The auxiliary encoder shares the identical architecture and weight maxtrix with the comparative classifier (CNN), and the latent dimension is set to 100.

\paragraph{Optimization.} We use Adam optimizer with $\beta_1=0.9$ and $\beta_2=0.999$. The initial learning rate for the generator was set to 1e-2 and 1e-4 for pretraining and adversarial training respectively. We set the initial learning rate as 1e-4 for the discriminator network during adversarial training. To prevent overfitting, we clip the gradients of trainable parameters whose L2 norm exceeds 5.

\paragraph{Training Settings.} We conduct experiments to finetune the following hyperparameters: agent number $N_g = \{1,2,3,4,5\}$, batch size of $\{32, 64, 128\}$, Gumbel-Softmax temperature $\tau \in \{ 1, 0.5, 0.1, 0.01. 0.001 \}$. The training step for the generator and discriminator network is set to $g=1$ and $d=5$, respectively. The generator is pretrained for 150 epochs before the adversarial training. Finally, the optimal batch size is set to 128 for both synthetic and real datasets. It is worth noting that we also test the batch size to 256, which requires too much GPU resource but do not show obvious improvement. All experiments are run with five different random seeds on single Nvidia Titan RTX GPU.

\begin{figure*}[h]
        \centering
        \begin{subfigure}[b]{0.475\textwidth}
            \centering
            \includegraphics[width=\textwidth]{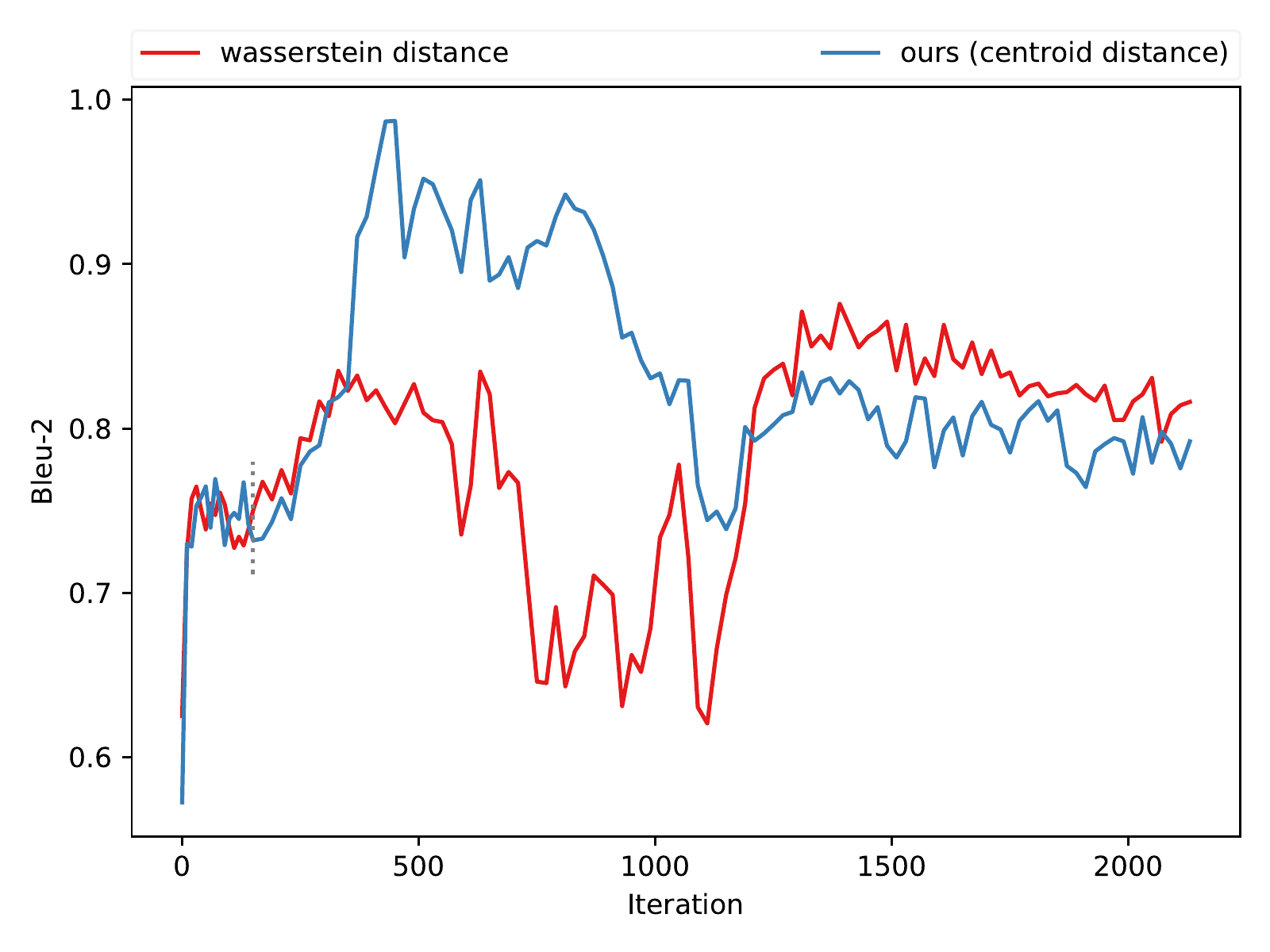} 
            \label{fig:metric_B2}
        \end{subfigure}
        \hfill
        \begin{subfigure}[b]{0.475\textwidth}  
            \centering 
            \includegraphics[width=\textwidth]{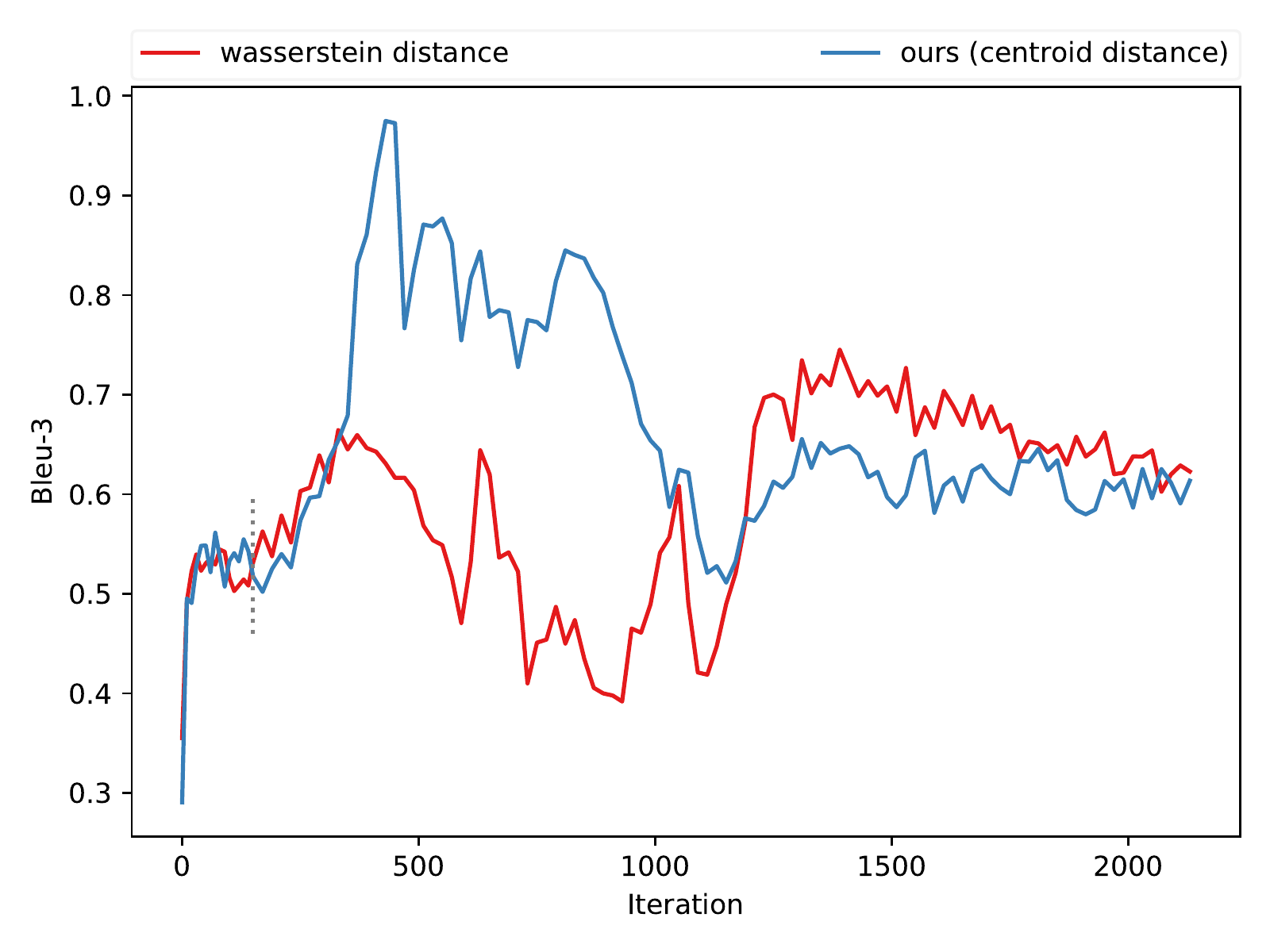} 
            \label{fig:metric_B3}
        \end{subfigure}
        \vskip\baselineskip
        \begin{subfigure}[b]{0.475\textwidth}   
            \centering 
            \includegraphics[width=\textwidth]{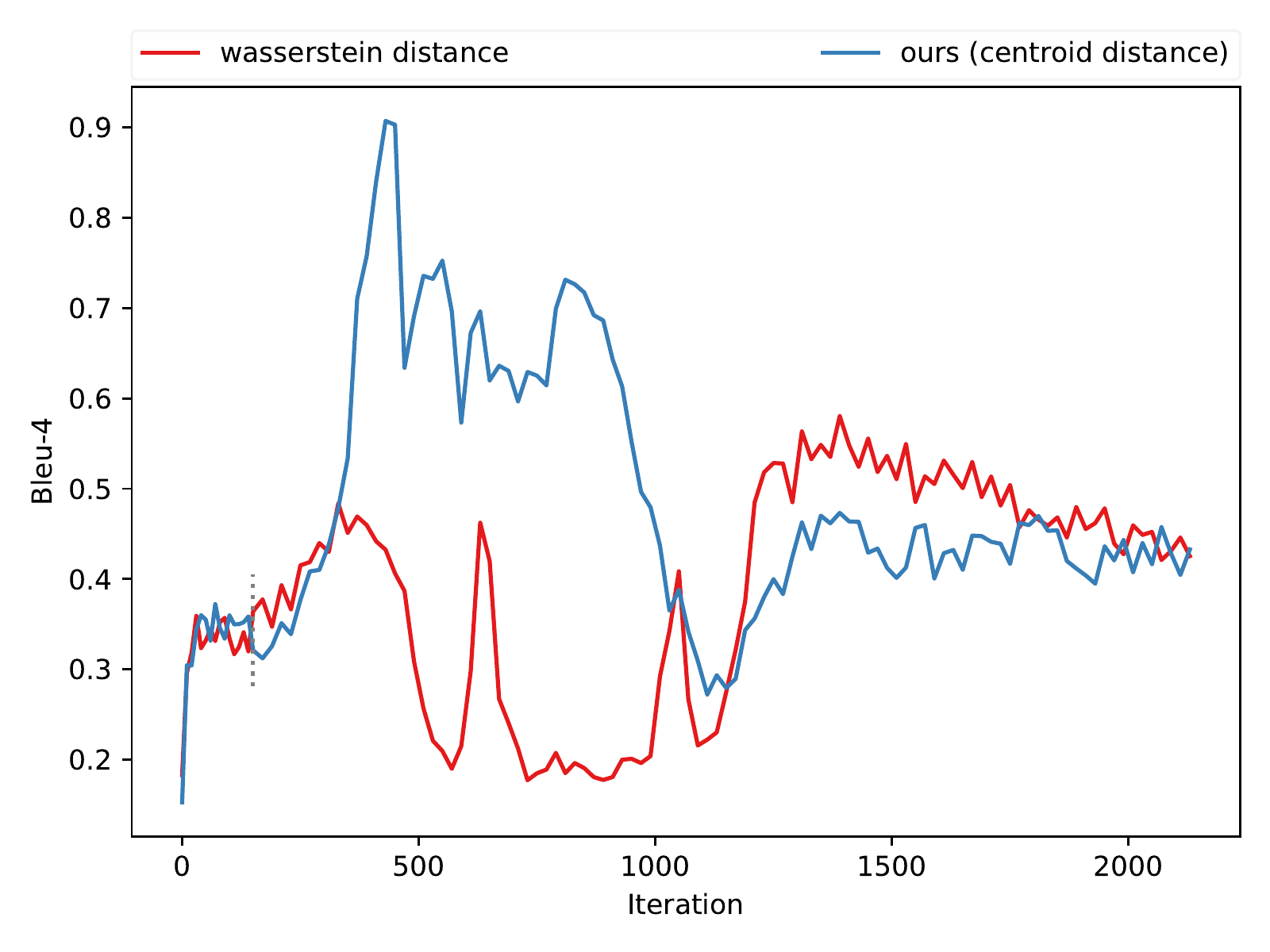}
            \label{fig:metric_B4}
        \end{subfigure}
        \hfill
        \begin{subfigure}[b]{0.475\textwidth}   
            \centering 
            \includegraphics[width=\textwidth]{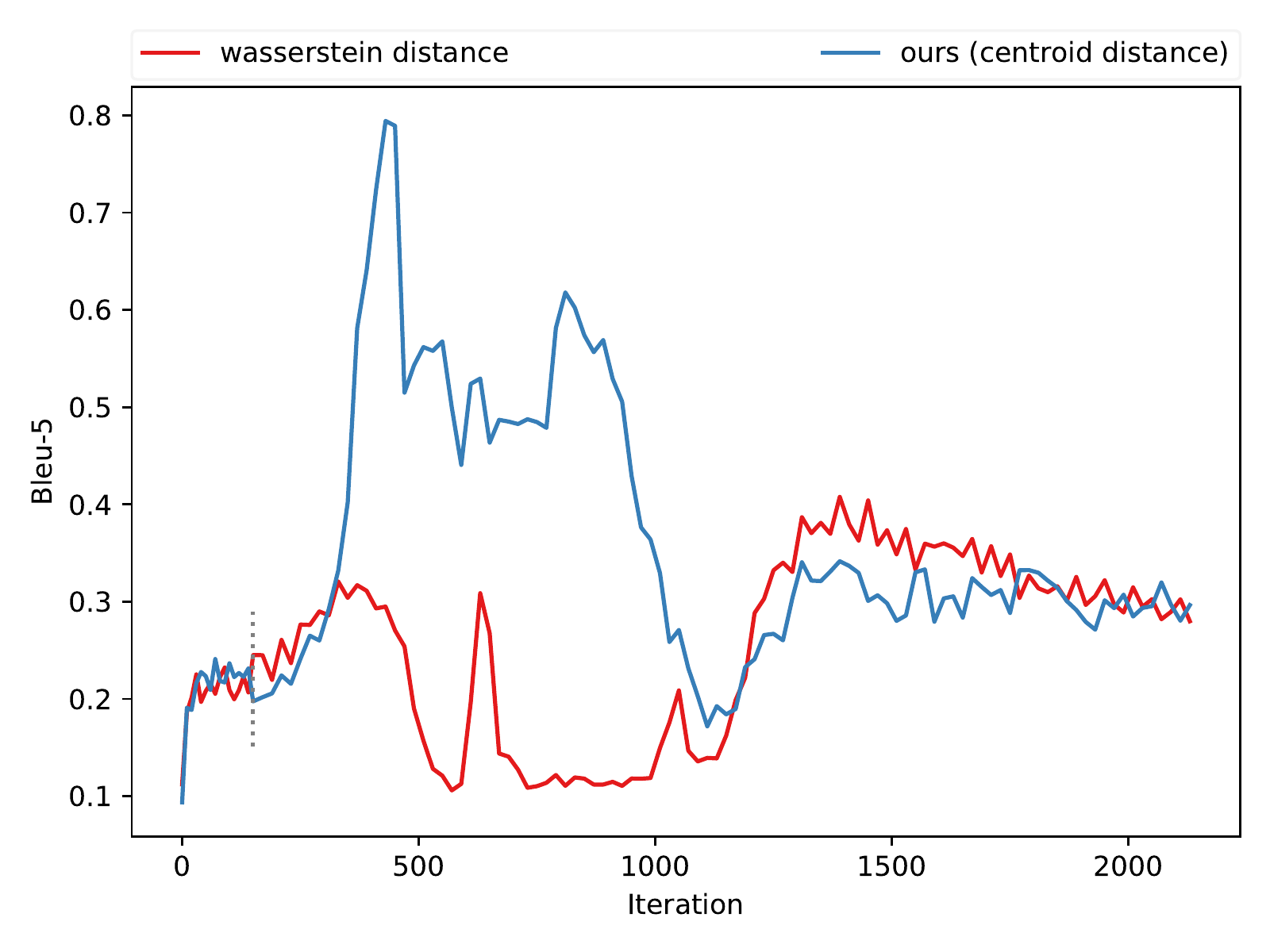}
            \label{fig:metric_B5}
        \end{subfigure}  
         \begin{subfigure}[b]{0.475\textwidth}   
            \centering 
            \includegraphics[width=\textwidth]{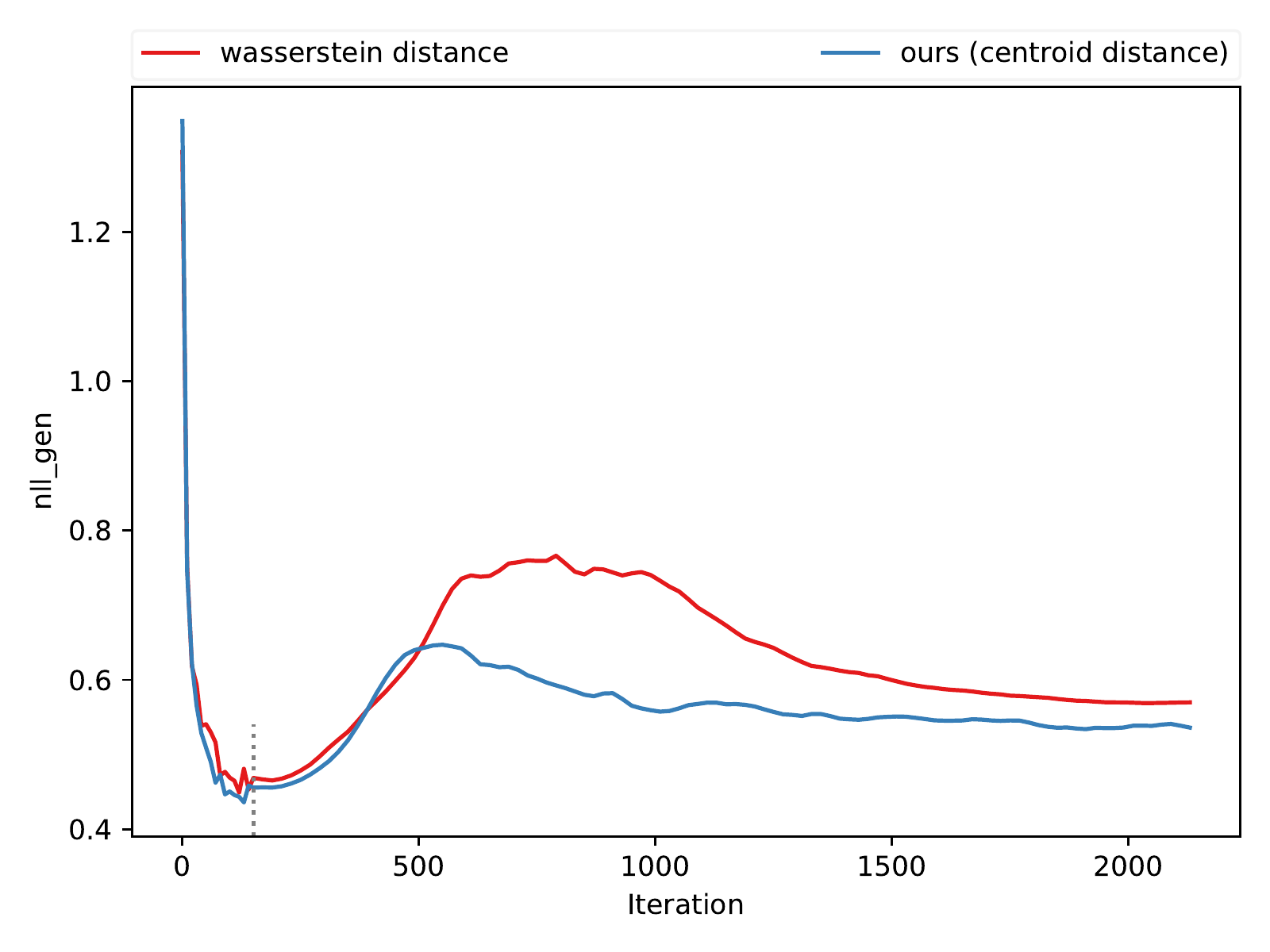}
            \label{fig:metric_nll_gen}
        \end{subfigure}
        \caption[  ]
        {\small Training curves of BLEU scores on MS COCO Image Caption dataset with the Wasserstein distance and our FSA distance (centroid distance). The vertical dash lines indicate the end of generator pretraining. For BLEU scores, the higher, the better.} 
        \label{fig:metric_BLEU}
    \end{figure*}
    
\section{Negative Results}
\label{ap:neg_results}
We list some approaches we tried but proved unsuccessful:
\begin{itemize}
  \item Replacing the FSA distance with Wasserstein distance to verify the effect of different distance for Feature Statistics Alignment. We found that the Wasserstein distance is not as efficient as our method that measures the distance between real and fake distribution centroids (See Figure~\ref{fig:metric_BLEU} for comparison performance).
  \item Using Mogrifier LSTM as the generator, which achieves similar results as vanilla LSTMs on the synthetic data. 
  \item Using a Wasserstein loss instead of current Relativistic Discriminator. Not as stable as the current solution. 
  \item Using the Transformer model as the discriminator. It achieves unsatisfied results with the current experimental settings.
  \item Using interleaved training instead of two-stage training, \emph{i.e.,} adversarial training after pretraining. It is unsuccessful to train the generator for 15 iterations after one iteration using MLE.
  \item Using top-k sampling and nucleus sampling, instead of the argmax in the Gumbel-Max trick. This does not always boost the final performance.
  \item Using a hinge loss on the discriminator. This did not improve over the current relativistic loss.
\end{itemize}

\section{Evaluation Details}
\subsection{Synthetic Data}
\label{ap:synthetic}
For synthetic data, we evaluate the generated sequence w.r.t. both quality and diversity. We use the oracle LSTM to evaluate the negative log-likelihood of our generated samples (denoted as NLL\textsubscript{oracle}) to measure the quality, and the negative log-likelihood of the synthetic dataset (denoted as NLL\textsubscript{gen}) measured by the generator during training. We also report the best NLL\textsubscript{oracle}+NLL\textsubscript{gen} to evaluate the trade-off between quality and diversity. It is observed that our model outperforms baseline models in terms of quality (measured by NLL\textsubscript{oracle}) and quality-diversity trade-off (measured by NLL\textsubscript{oracle}+NLL\textsubscript{gen}), and achieves or matches the competitive results of baselines w.r.t. the diversity (indicated by NLL\textsubscript{gen}).

\begin{table}[thb]
\centering
\resizebox{\linewidth}{!}{%
\begin{tabular}{@{}l|ccc@{}}
\toprule
Model   & NLL\textsubscript{oracle} (20/40)                 & NLL\textsubscript{gen} (20/40)                     & NLL\textsubscript{oracle} + NLL\textsubscript{gen} (20/40) \\ \midrule
MLE  & 9.05{\small$\pm$0.03} / 9.84{\small$\pm$0.02} & 5.96{\small$\pm$0.02} / 6.55{\small$\pm$0.02} & 15.02{\small$\pm$0.03} / 16.39{\small$\pm$0.01}   \\
SeqGAN & 8.63{\small$\pm$0.19} / 9.63{\small$\pm$0.04} & 6.61{\small$\pm$0.22} / 6.98{\small$\pm$0.08} & 15.00{\small$\pm$0.03} / 16.35{\small$\pm$0.02}   \\
RankGAN & 8.42{\small$\pm$0.31} / 9.52{\small$\pm$0.11} & 7.14{\small$\pm$0.34} / 7.05{\small$\pm$0.12} & 15.01{\small$\pm$0.02} / 16.37{\small$\pm$0.02}   \\
MaliGAN & 8.74{\small$\pm$0.16} / 9.67{\small$\pm$0.03} & 6.62{\small$\pm$0.25} / 7.14{\small$\pm$0.09} & 15.03{\small$\pm$0.03} / 16.39{\small$\pm$0.03}   \\
RelGAN & 6.73{\small$\pm$0.54} / 6.68{\small$\pm$0.20} & 6.38{\small$\pm$0.70} / 7.17{\small$\pm$0.69} & 13.11{\small$\pm$0.55} / 13.85{\small$\pm$0.54}   \\
SAL   & 7.71{\small$\pm$0.17} / 9.31{\small$\pm$0.03} & 6.58{\small$\pm$0.15} / 6.97{\small$\pm$0.05} & 14.29{\small$\pm$0.11} / 16.24{\small$\pm$0.03}   \\ 
CCL    & 6.77{\small$\pm$0.34} / 6.65{\small$\pm$0.14} & 6.91{\small$\pm$0.62} / 7.68{\small$\pm$0.79} & 13.69{\small$\pm$0.36} / 14.33{\small$\pm$0.76}\\
\midrule
Ours    & \textbf{5.84}{\small$\pm$0.35} / \textbf{5.71}{\small$\pm$0.29} & \textbf{5.07}{\small$\pm$0.78} / 7.46{\small$\pm$0.66} & \textbf{10.90}{\small$\pm$0.59} / \textbf{10.17}{\small$\pm$0.43}   \\ \bottomrule
\end{tabular}%
}
\caption{Performance of different models on the synthetic dataset with the sequence length of 20 and 40, respectively. For NLL scores, the lower, the better.}
\label{tab:syn_all}
\end{table}

\subsection{Human Evaluation}
\label{ap:human_eval}
Acceptance (\emph{i.e.} whether a sentence is acceptable by human beings), grammaticality (\emph{i.e.}, if a sentence is grammatically correct), and meaningfulness (\emph{i.e.}, if a sentence makes sense) are three main standards for the text quality evaluation. Please note that any minor text formatting issues which will not negatively influence the understanding and correctness of the sentences (\emph{e.g.}, punctuation, capitalization, spelling errors, extra spaces) can be ignored. Please also note: a sentence consists of less than 10 words should get one point deducted. Table~\ref{tab:human_eval_criteria} gives more detailed criteria.  

It is worth mentioning that human evaluation is used to measure the quality of generated sentences rather than diversity. For comprehensive comparison, we also compare with samples of MaliGAN~\cite{Che2017MaximumLikelihoodAD} and TextGAN~\cite{Zhang2017AdversarialFM} in addition to aforementioned baselines.

\begin{table*}[ht]
\centering

\resizebox{.9\textwidth}{!}{%
\begin{tabular}{@{}ll@{}}
\toprule
Scale & Criterion \& Example \\ \midrule
5 - Excellent & \begin{tabular}[c]{@{}l@{}}Grammatical, acceptable, and meaningful. For example, ``a man is carving under yellow\\ planes .''\end{tabular} \\
4 – Good & \begin{tabular}[c]{@{}l@{}}Include 1 to 2 tiny grammatical errors, and the whole sentence is mostly acceptable and \\ meaningful. For example, ``two giraffe   standing in front of them .''\end{tabular} \\
3 – Fair & \begin{tabular}[c]{@{}l@{}}Include major grammatical errors, but the whole sentence is still acceptable and making  \\ sense. For example, ``a kitchen with a grill roll from him .''\end{tabular} \\
2 – Poor & \begin{tabular}[c]{@{}l@{}}Include severe grammatical errors. The whole sentence does not make sense,  but some \\ parts are still acceptable. For example, ``a motorcycle on a   paved road on the freeway .''\end{tabular} \\
1 - Unacceptable & \begin{tabular}[c]{@{}l@{}}It is basically a string of words with random order and totally ungrammatical. The entire  \\ sentence does not make any sense. For example, ``a city .''\end{tabular} \\ \bottomrule
\end{tabular}%
}
\caption{The human evaluation scale from 1 to 5 with corresponding criteria and example sentences.}
\label{tab:human_eval_criteria}
\end{table*}

\begin{figure*}[t]
        \centering
        \begin{subfigure}[b]{0.475\textwidth}
            \centering
            \includegraphics[width=\textwidth]{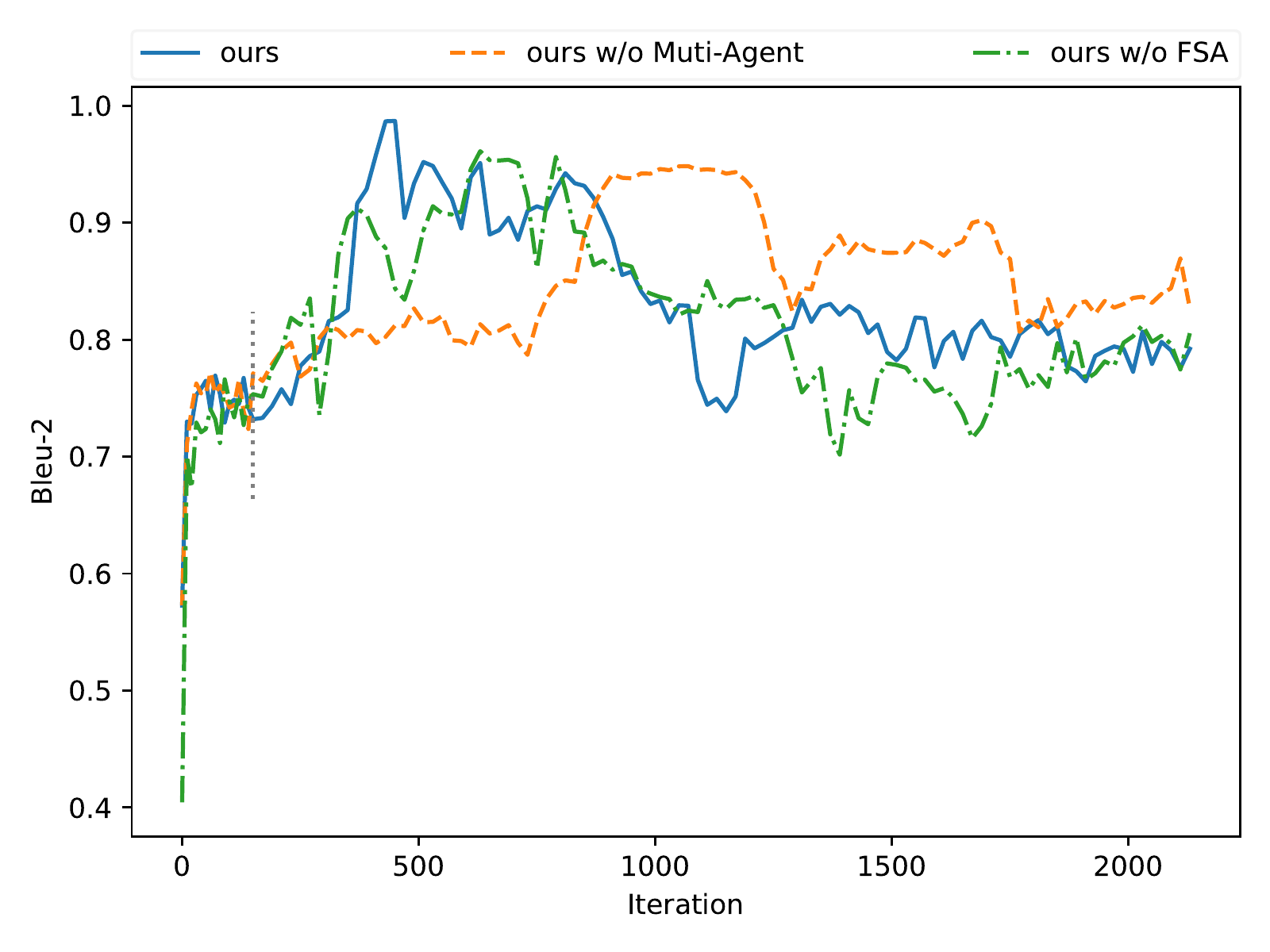} 
            \label{fig:abl_B2}
        \end{subfigure}
        \hfill
        \begin{subfigure}[b]{0.475\textwidth}  
            \centering 
            \includegraphics[width=\textwidth]{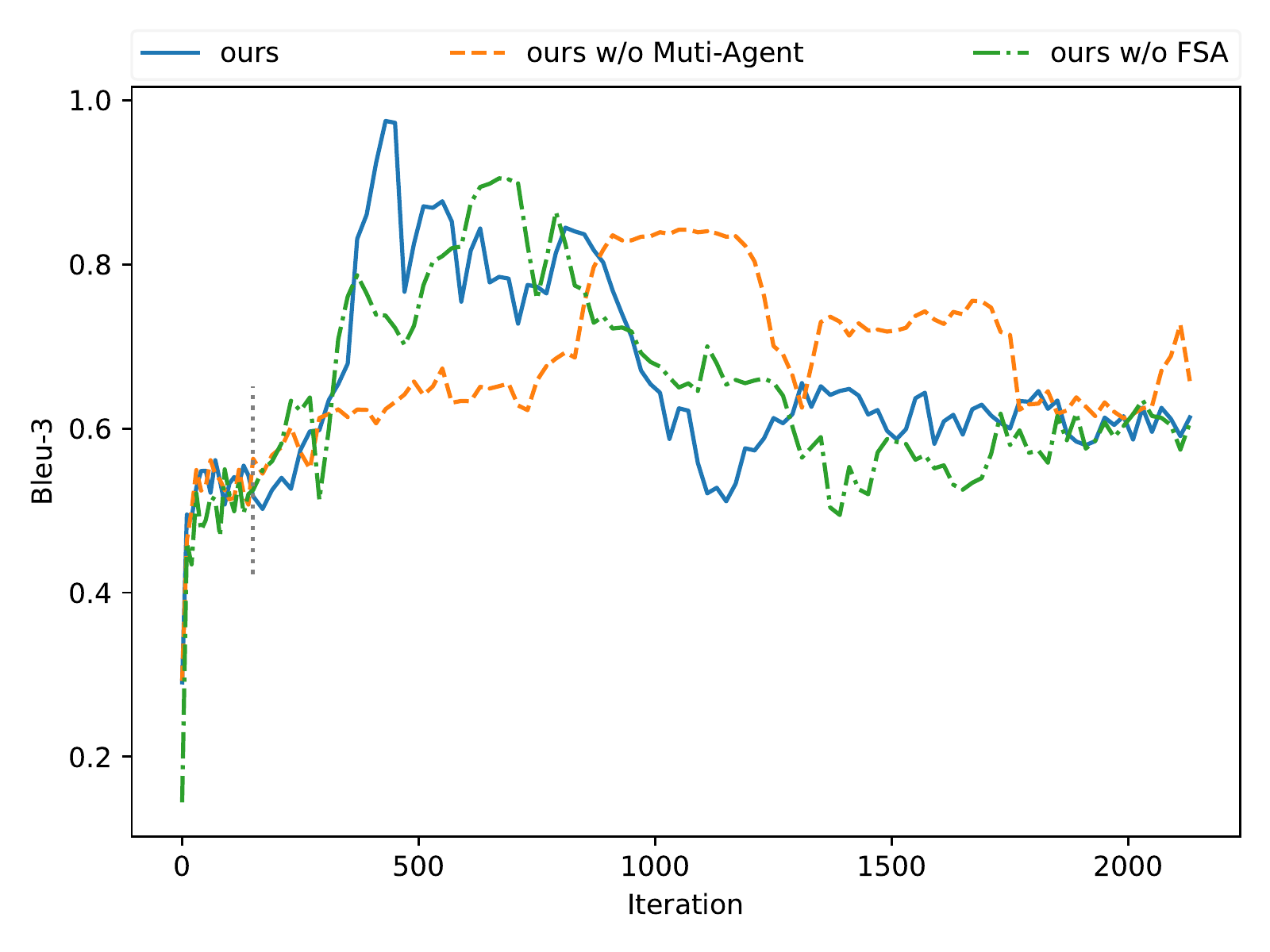} 
            \label{fig:abl_B3}
        \end{subfigure}
        \vskip\baselineskip
        \begin{subfigure}[b]{0.475\textwidth}   
            \centering 
            \includegraphics[width=\textwidth]{fig/ablation_Bleu-4.pdf}
            \label{fig:abl_B4}
        \end{subfigure}
        \hfill
        \begin{subfigure}[b]{0.475\textwidth}   
            \centering 
            \includegraphics[width=\textwidth]{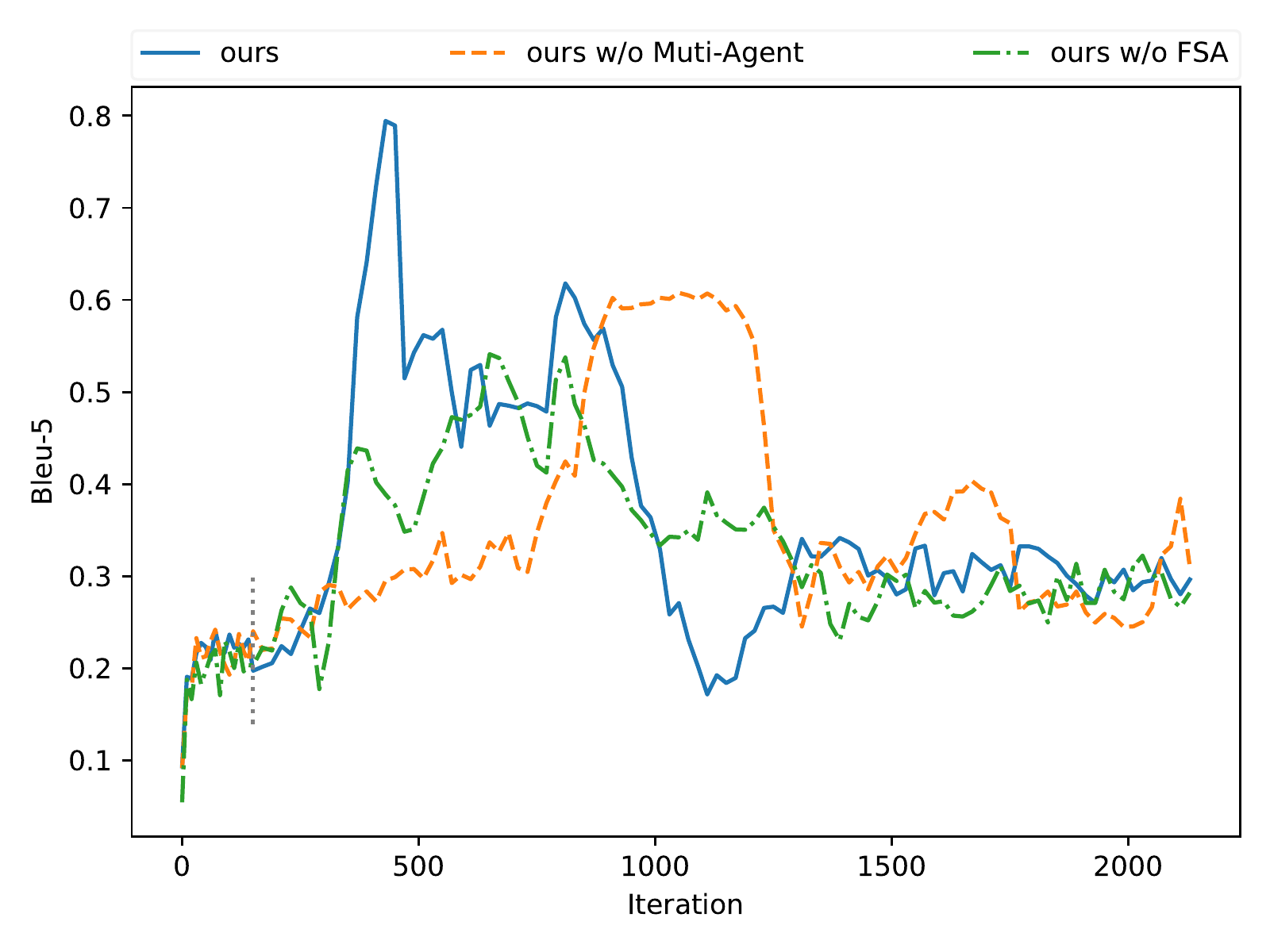}
            \label{fig:abl_B5}
        \end{subfigure}  
                \begin{subfigure}[b]{0.475\textwidth}   
            \centering 
            \includegraphics[width=\textwidth]{fig/ablation_nll_gen.pdf}
            \label{fig:abl_nll_gen}
        \end{subfigure} 
        \caption[  ]
        {\small Training curves of BLEU scores on MS COCO Image Caption dataset w/ and w/o mixture-of-experts generator and FSA mechanism. The vertical dash lines indicate the end of generator pretraining. For BLEU scores, the higher, the better.} 
        \label{fig:ablation_BLEU}
    \end{figure*}

\subsection{Linguistic Analysis}

From the syntactic perspective, our model is well performed by generating grammatically correct and meaningful sentences in most instances. Great deals of the generated sentences follow the basic sentence pattern in English, which is SVO (Subject Verb Object), with few exceptions. One of the most common ungrammatical forms is the omission of the verb. For example, in the sentence ``a man standing next to a small airplane with two dogs'' the primary verb ``is'' did not appear. Hence, samples like this cannot be counted as grammatically acceptable sentences but only meaningful phrases. However, even though there is a grammatical error, the sentences mostly make sense, which can be counted in the scale of 4. This is because there is plenty of sentences omitting verbs in real training data, such as ``a sink next to a large white door'' and ``city street with parked cars and a bench''. It thus makes sense that models generate meaningful sentences with predicate omission.

Another point worth noting is the syntactic ambiguity, resulting from the placement of preposition phrases. For instance, the sentence ``a man is sitting on a motorcycle with a woman on the bike'' can have at least two kinds of interpretations. One of the explanations could be – a man is sitting on a motorcycle, while a woman is sitting on a bike. Another absurd meaning could be – a man and a woman are riding a motorcycle while sitting on the bike. The example shows that even the model can generate grammatically correct sentences, not all meanings make sense. Therefore, it still remains challenges to generate sentences with better syntactic meaning.

\subsection{Case Study}
\label{ap:case}
Table~\ref{tab:samples} displays the generated samples from all baseline models and references in the MS COCO Image Caption dataset. From the presented sentences, we can observe that samples generated by MLE is less meaningful than other models, which is consistent with the results in~\cite{Zhou2020SelfAdversarialLW}. Besides, models such as RankGAN, RelGAN, and ours, tend to produce realistic sentences. RelGAN tends to generate long sentences with more prepositional phrases but lacks of consistency of language context, such as ``with a chair on the couch''. SAL also generates confusing words such as ``a man on a motorcycle is flying''. In contrast, our method can generate prepositional phrases more appropriately, such as ``on a beach next to the ocean''. We can observe that our method generates better human-looking samples than other GANs and the MLE baseline.

\begin{table*}[h]
\centering

\resizebox{\linewidth}{!}{%
\begin{tabular}{@{}l|l@{}}
\toprule
          & Samples                                                                                                                                                                                                         \\ \midrule
Real       & \begin{tabular}[c]{@{}l@{}}a man flies a kite in a park .\\ a man riding a bike with a wooden trainer attached and a dog riding in it .\end{tabular}     \\ \midrule
MLE        & \begin{tabular}[c]{@{}l@{}}a man watches on his bike , in a lake on a field .\\ a women is standing behind an orange table in helmet on a child in the background .\end{tabular}                                \\ \midrule
SeqGAN     & \begin{tabular}[c]{@{}l@{}}some people sitting on top of luggage near a truck .\\ a man sitting in a bath tub on tops .\end{tabular}                                                                            \\ \midrule
TextGAN    & \begin{tabular}[c]{@{}l@{}}a man riding a motorcycle .\\ a bathroom with a sink , and a table .\end{tabular}                                                                                                    \\ \midrule
LeakGAN    & \begin{tabular}[c]{@{}l@{}}a man standing next to her cell phone on a street sign .\\ a woman is holding a child in the air .\end{tabular}                                                                      \\ \midrule
MaliGAN    & \begin{tabular}[c]{@{}l@{}}a woman is standing and another oak cake on a drain .\\ a man standing in a kitchen with her laptop and two tables\end{tabular}                                                      \\ \midrule
RankGAN    & \begin{tabular}[c]{@{}l@{}}a colorful bike is is down next to a large mirror .\\ a man is riding a bike down a track .\end{tabular}                                                                             \\ \midrule
RelGAN     & \begin{tabular}[c]{@{}l@{}}a woman walking with a dog in the city in front of a city bus .\\ a man sitting on a bed in a room with a chair on the couch .\end{tabular}                                          \\ \midrule
SAL     & \begin{tabular}[c]{@{}l@{}}
a man on a motorcycle is flying on a grassy field .\\ a man stands in a green field .\end{tabular}                                          \\ \midrule
Ours & \begin{tabular}[c]{@{}l@{}} a person is flying a kite on a beach next to the ocean .\\  a man is cooking in a kitchen with a white toilet in the background .\end{tabular}    \\ \bottomrule
\end{tabular}%
}
\caption{Samples of baseline models and real data on MS COCO Image Caption dataset.}
\label{tab:samples}
\end{table*}

\begin{table*}[thb]
\centering
\resizebox{.6\textwidth}{!}{%
\begin{tabular}{@{}l@{}}
\toprule
\begin{tabular}[c]{@{}l@{}}a person is flying a kite on a beach next to the ocean .\\\\ a large jet sitting on a runway next to a landing strip .\\\\ a bathroom with a toilet , a sink , and towels hanging on a rack . \\\\ a man is sitting on a motorcycle with a woman on the back . \\\\ a man is cooking in a kitchen with a white toilet in the background . \\\\ a man is sitting a motorcycle on a dirt road . \\\\ an airplane is flying high in the sky .\\\\ a man and a woman on a motorcycle in front of a building .\\\\ a herd of sheep grazing in a pasture .\\\\ a cluttered room has large bed and a large clock .\end{tabular} \\ \bottomrule
\end{tabular}%
}
\caption{Randomly sampled 10 samples trained on MS COCO Image Captions.}
\label{tab:samples_coco}
\end{table*}

\begin{table*}[thb]
\centering
\resizebox{\textwidth}{!}{%
\begin{tabular}{@{}l@{}}
\toprule
\begin{tabular}[c]{@{}l@{}}when it was announced on february 10 , it was delivered to ensure that everyone involved will cut operating costs .\\\\ but it was always clear that we could see what it is about and what ' s going on in the next round . \\\\ even though economists have not yet been able to speak out after the election result , she insists . \\\\ ``but it ' s a natural question for ordinary people in the uk and across the uk . '' he said .\\\\ but it is still only six months until now , according to a new survey .\\\\ and it was a real surprise for us , and now i ' m optimistic , happy in the next couple of weeks .\\\\ ``but it is a signal that we are carrying out our own power in the wake of this period , '' he said . \\\\ russia is its only one in the top of the list since the arrest warrant launched an operation on its own behalf .\\\\ it is therefore no evidence that the person who committed to office is involved in the operation , but no one has to \\say any seriously wounded .\\\\ ``but it was a very poor start for me and i ' d imagine that ' s what we ' re doing and we ' re looking at focusing \\on any given contract , '' he said .\end{tabular} \\ \bottomrule
\end{tabular}%
}
\caption{Randomly sampled 10 samples trained on EMNLP2017 WMT News.}
\label{tab:samples_emnlp}
\end{table*}

\section{Results of Ablation Study}
\label{ap:ablation}
Figure~\ref{fig:ablation_BLEU} presents all BLEU scores and NLL\textsubscript{gen} of the ablation study. Our model enjoys the advantage of both the mixture-of-experts and FSA paradigm, achieving the superior performance on both the quality and diversity metrics. This demonstrates the effectiveness of proposed methods on improving the quality of generated samples. It is observed that our full model can reach its peak more quickly than models with any ablations on the mixture-of-experts generator or the FSA method. Removing either of them can deteriorate the overall performance on all BLEU scores and NLL\textsubscript{gen} metric. It can be inferred that the FSA method can have positive effects on the sample diversity while the mixture-of-experts generator can enrich the representation capacity of the generator and thus promote the training efficiency of the generator network.

\section{Generated Samples on Real Data}
\label{ap:sample}

Table~\ref{tab:samples_coco} and \ref{tab:samples_emnlp} show the randomly sampled sentences from the proposed models generated on MS COCO Image Captions and EMNLP2017 WMT News dataset, respectively. We can see that the proposed method can generate meaningful and grammatically correct sentences on real data.

\end{document}